\documentclass[final,5p,times,twocolumn]{elsarticle}

\usepackage{amssymb}
\usepackage{amsmath}

\usepackage{amsmath,amsfonts,bm}

\def\eqref#1{equation~\ref{#1}}

\def\1{\bm{1}}

\def\vh{{\bm{h}}}

\def\vw{{\bm{w}}}
\def\vx{{\bm{x}}}

\def\mI{{\bm{I}}}

\def\mW{{\bm{W}}}

\def\mZ{{\bm{Z}}}

\DeclareMathAlphabet{\mathsfit}{\encodingdefault}{\sfdefault}{m}{sl}
\SetMathAlphabet{\mathsfit}{bold}{\encodingdefault}{\sfdefault}{bx}{n}

\DeclareMathOperator*{\argmax}{arg\,max}

\usepackage{graphicx}
\usepackage{booktabs}
\usepackage{hyperref}
\usepackage{multirow}
\usepackage{multicol}
\usepackage{float, booktabs, boldline, hhline}
\usepackage{wrapfig}
\usepackage{soul}
\usepackage{xcolor}
\usepackage{color, colortbl}
\usepackage{subcaption}
\usepackage{enumitem}
\definecolor{tabhighlight}{HTML}{e5e5e5}
\usepackage{tabularx}
\usepackage{bbding}
\usepackage{balance}

\journal{Pattern Recognition}
\makeatletter
\def\ps@pprintTitle{%
    \let\@oddhead\@empty
    \let\@evenhead\@empty
    \let\@oddfoot\@empty
    \let\@evenfoot\@empty
}
\makeatother

\begin{document}

\begin{frontmatter}



\title{Cluster-Aware Prompt Ensemble Learning for Few-Shot
Vision–Language Model Adaptation}


\author[1]{Zhi Chen}
\author[2]{Xin Yu}
\author[1]{Xiaohui Tao}
\author[1]{Yan Li}
\author[2]{Zi Huang} 

\affiliation[1]{organization={University of Southern Queensland},
            city={Toowoomba},
            postcode={4350}, 
            state={Queensland},
            country={Australia}}
\affiliation[2]{organization={University of Queensland},
            city={Brisbane},
            postcode={4072}, 
            state={Queensland},
            country={Australia}}
\begin{abstract}

Vision-language models (VLMs) such as CLIP achieve zero-shot transfer across various tasks by pre-training on numerous image-text pairs. These models often benefit from using an ensemble of context prompts to represent a class. Despite being effective, conventional prompt ensembling that averages textual features of context prompts often yields suboptimal results. This is because feature averaging shifts the class centroids away from the true class distribution. To address this issue, we propose the Cluster-Aware Prompt Ensemble Learning (CAPEL) framework, which preserves the cluster nature of context prompts.
CAPEL classifies images into one of several class clusters, each represented by a distinct prompt. Instead of ensembling prompts in the feature space, we perform ensembling in the classification logits space, aligning better with the visual feature distribution. To further optimize prompt fine-tuning while maintaining cluster-specific discriminative power, we introduce a cluster-preserving regularization term. This ensures that prompts remain distinct and specialized for different clusters, preventing collapse into a uniform direction. Additionally, we integrate an adaptive prompt weighting technique to dynamically adjust the attention weights for flawed or ambiguous prompts, ensuring robust performance across diverse datasets and tasks.

\end{abstract}







\end{frontmatter}



\section{Introduction}
\label{sec:intro}

Recent research in pre-trained Vision-Language Models (VLMs), such as CLIP \cite{radford2021learning} and ALIGN \cite{jia2021scaling}, have demonstrated remarkable capabilities in zero-shot recognition. These models can classify previously unseen images into categories without explicitly training to recognize them. To achieve strong zero-shot classification performance, prompt engineering \cite{radford2021learning,pham2023combined} is usually required. For instance, adding the prompt \textit{``A photo of a \{\}."} boosts CLIP ViT-B/16 ImageNet classification accuracy from 64.18\% to 66.92\%. An ensemble of 80 hand-crafted prompts further improves accuracy to 68.57\% \cite{radford2021learning}. In such cases, the textual features of prompts are averaged to represent a single class vector, which enhances zero-shot performance. 

\begin{figure}[t]
\begin{center}
\vspace{-10pt}
\includegraphics[width=0.9\linewidth]{ 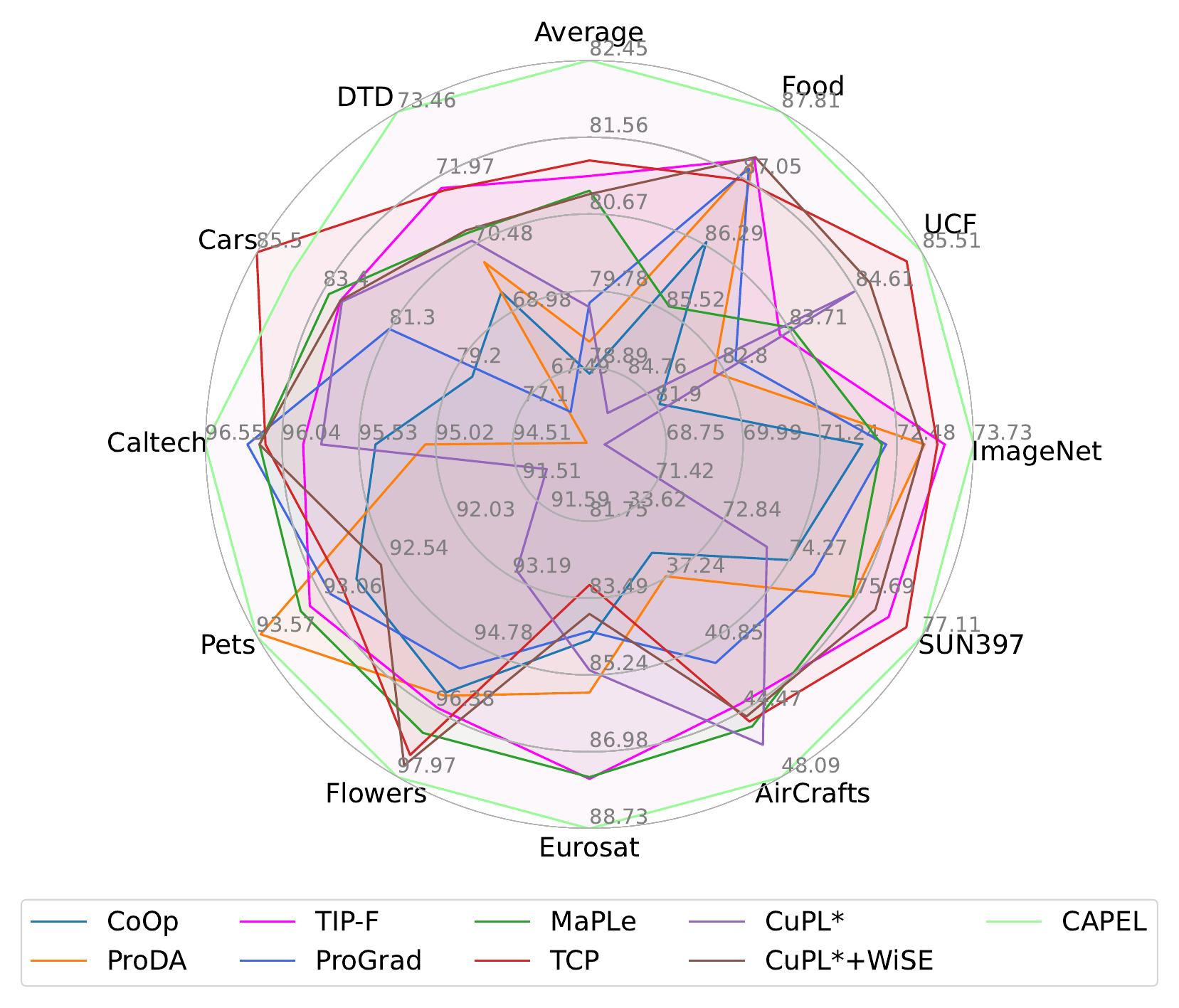}
\vspace{-20pt}
\end{center}
\caption{Performance comparison with state-of-the-art few-shot adaptation methods in 16-shot setting on 11 datasets. Our proposed CAPEL consistently achieves competitive performance.}
\label{performance}
\vspace{-10pt}
\end{figure}

Following prompt ensembling methods such as CuPL \cite{pratt2023does} refine this process by using large language models to generate more contextually (class names) relevant prompts. This aligns better with VLMs' training data, where class names are often paired with descriptive words. However, averaging textual features has its limitations. 
To address this, methods like WiSE \cite{wortsman2022robust} propose combining zero-shot and fine-tuned classifiers to improve robustness. Another research line such as ZPE \cite{allingham2023simple}, is to perform a weighted averaging by considering the importance of different prompts.

\begin{figure*}[t]
\begin{center}
\includegraphics[width=0.99\linewidth]{ 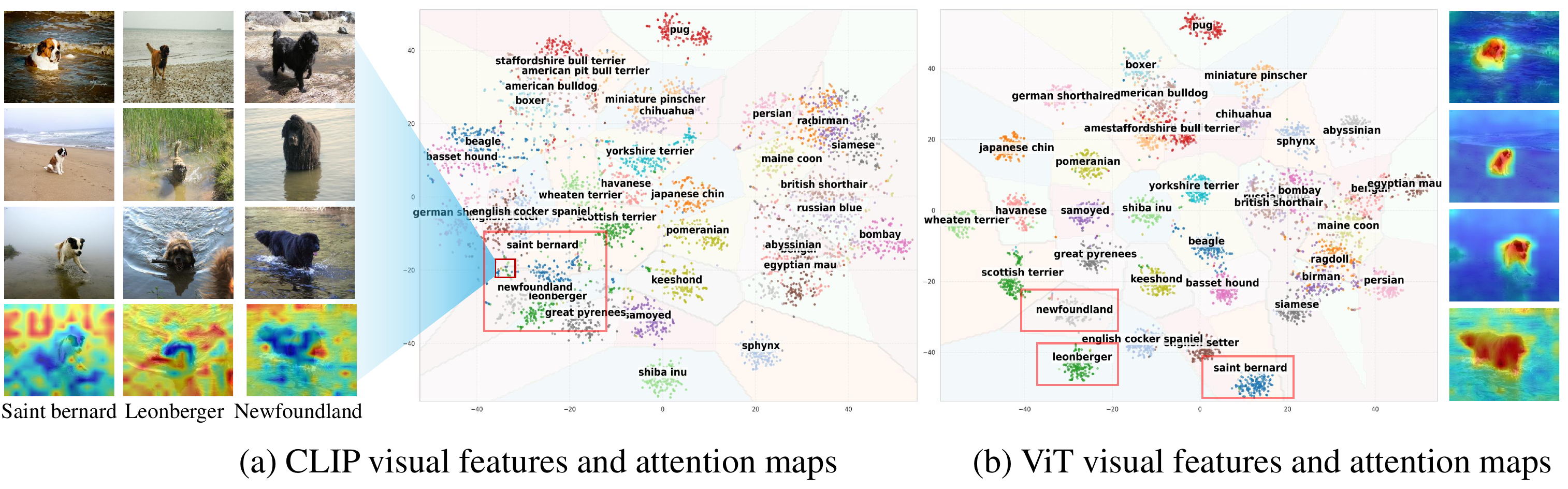}
\end{center}
\vspace{-10pt}
\caption{\textbf{\textit{Visualization of Visual Feature Clusters in the Oxford Pets Dataset.}} (a) CLIP visual features are semantically rich: This plot represents the visual feature clustering generated by the vision-language model CLIP. Since CLIP is pre-trained on diverse image-text pairs, it captures contextual information in the background as well as the primary object. For example, a small cluster contains \textbf{water-related backgrounds} associated with breeds like Saint Bernard, Leonberger, and Newfoundland, indicating that CLIP considers both the foreground (pet) and background context. (b) Vision Transformer (ViT) Encoder: This plot shows feature clustering using a ViT model pre-trained on image-label pairs. Unlike CLIP, ViT is more focused on the primary object itself, which results in clusters that are more distinctly separated by breed type without background influence.}
\label{pets_vis}
\end{figure*}

\begin{figure}[t]
    \centering
    \includegraphics[width=0.99\linewidth]{ 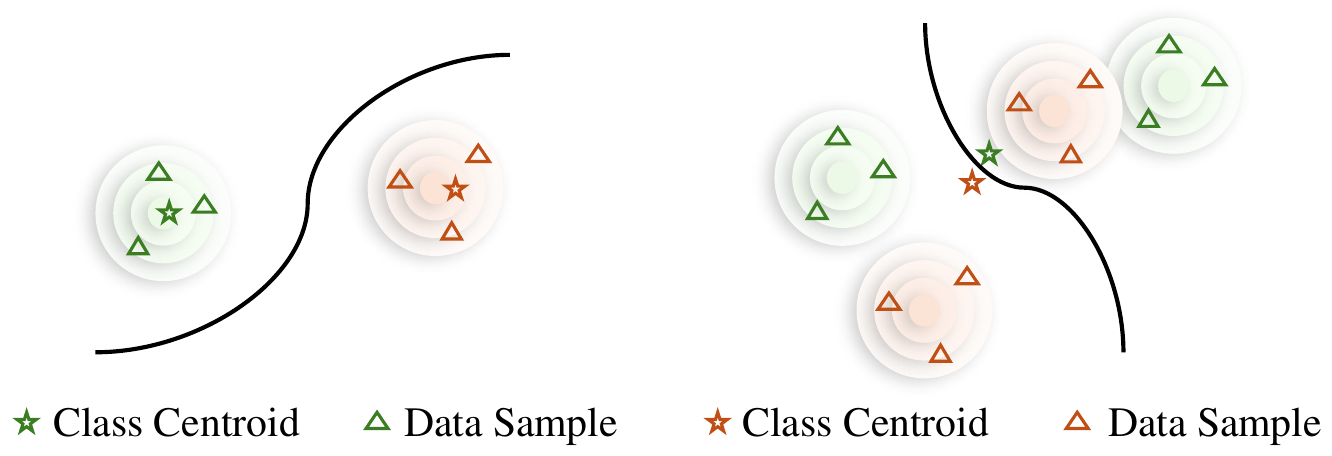}
    \caption{Local visualization of feature spaces. \textbf{Left}: Supervised vision backbones produce tightly clustered features within classes, with class centroids (stars) located at the center of the clusters. \textbf{Right}: Vision-Language Models (VLMs) capture contextual information across categories, forming multiple sub-clusters for each class, often displacing the centroids (stars) outside the main clusters. This difference highlights the richer contextual embeddings of VLMs compared to traditional supervised models.}
    \label{local_vis}
\end{figure}

While effective, these approaches operate in the feature space, which may not fully capture the multi-cluster nature of visual data.  As shown in Fig. \ref{pets_vis}, our observations indicate that VLMs like CLIP exhibit strong sensitivity to background information due to their paired image-text training. For example, while Vision Transformers (ViT) focus primarily on foreground objects, CLIP's attention often includes background-context associations. 
This tendency results in class clusters influenced by specific contextual information, such as water-related background. As shown in Fig. \ref{local_vis}, directly averaging textual features can misalign these clusters, shifting class centroids and producing suboptimal decision boundaries, which may degrade performance when fine-tuned for downstream tasks. 

\begin{figure*}[t]
\begin{center}
\includegraphics[width=0.99\linewidth]{ 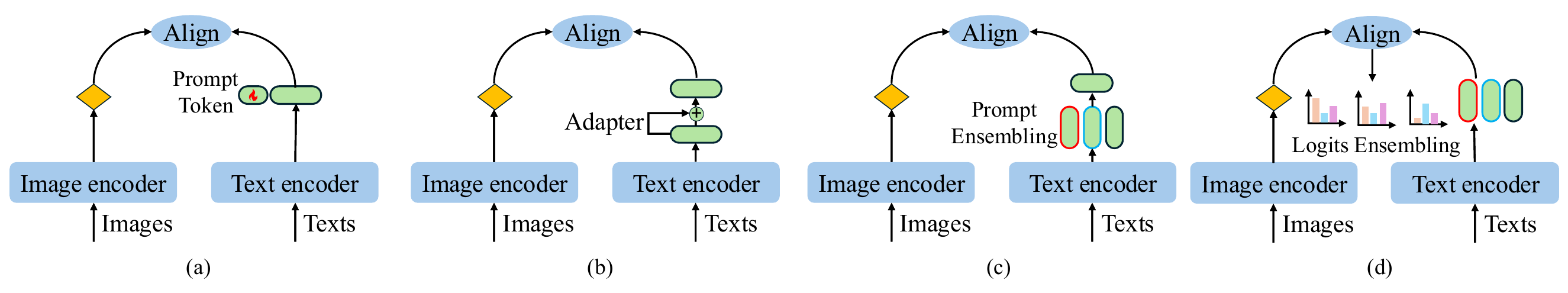}
\end{center}
\caption{Few-shot adaptation for VLMs. (a) Prompt Tuning \cite{zhou2022learning}. (b) Adapter-based methods \cite{gao2024clip}. (c) Prompt Ensembling \cite{pratt2023does}. (d) Prompt Logits Ensembling.}
\label{paradigm}
\vspace{-10pt}
\end{figure*}

To address this issue, in this paper, we propose Cluster-Aware Prompt Ensemble Learning (CAPEL), a fine-tuning strategy that averages logits instead of text features. This design assigns distinct clusters to separate classifiers, better reflecting the visual feature space's inherent structure. 

We identify two challenges of prompt ensembling learning in the logits space. The first one is that unrelated prompts within a class converge to the same centroid, reducing the diversity of clusters. To counteract this, we introduce a conditional entropy minimization technique that maintains classifier diversity while preserving uncertainty. Our method ensures robust adaptation and leverages the full potential of diverse prompts, addressing limitations of existing feature-space-based ensembling approaches. 

The second challenge is that the generated prompts often contain ambiguous and flawed keywords, which leads to erroneous classifier initialization. Thus, we introduce a learnable weighting mechanism using an attention matrix. This mechanism is designed to dynamically adjust the confidence levels, reducing reliance on flawed prompts while increasing attention to more accurate and contextually representative prompts. We compare our methods with state-of-the-art methods for adapting CLIP for downstream tasks in Fig. \ref{paradigm}.

\vspace{-5pt}
\begin{itemize}
    \item We propose a Cluster-Aware Prompt Ensemble Learning (CAPEL) framework for adapting vision-language models to downstream object recognition datasets. Unlike existing prompt ensembling methods that average text features to represent a class, we integrate prediction logits to preserve the multi-cluster nature in CLIP visual space.
    \vspace{-3pt}
    \item To prevent the diverse prompts collapsing towards the same direction, we propose a cluster-preserving loss that enforces each prototype consistently specializes in a distinct subset of the class. In addition, an adaptive prompt weighting technique is proposed to mitigate the negative impact of the flawed prompts generated by large language models.
    \vspace{-5pt}
    \item As shown in Fig. \ref{performance}, we present comprehensive experiments on 11 datasets covering a diverse set of visual recognition tasks. Specifically, we conduct experiments on few-shot learning and domain generalization settings, and the results demonstrate the effectiveness of our method.
\end{itemize}

\section{Related Work}
\noindent \textbf{Vision Language Models.} ~~  Inspired by the success of large-scale language models like BERT \cite{devlin2018bert} and GPT series \cite{radford2019language,brown2020language} in NLP and early Zero-Shot Learning paradigms \cite{chen2020canzsl,chen2020rethinking,chen2021entropy,chen2021mitigating,chen2021semantics,su2022distinguishing,chen2022federated,chen2022gsmflow,chen2023zero,guo2024fine,guo2024element,chen2025svip}, researchers began pre-training VLMs on large datasets to then fine-tune them on downstream tasks \cite{su2019vl,tan2019lxmert}. 
Recent trends in the field lean towards a unified model for vision and language that can be jointly trained on multiple tasks. 
The CLIP \cite{radford2021learning} model learns visual and text features in a zero-shot manner by training on a large set of images paired with natural language descriptions. Similarly, ALIGN model \cite{jia2021scaling} pushes the boundary by scaling up the data and the model size, achieving state-of-the-art performance on multiple benchmarks. Florence \cite{yuan2021florence} further extends the representations to fine-grained objects, videos, and multiple modalities. SEGL \cite{yang2023effective} systematically studies how to leverage auxiliary visual pretraining tasks to help train end-to-end vision language models.
Although these VLMs have learned generalized representations for both vision and language, adapting to downstream tasks remains a challenging research problem. 
There have been many tailored methods proposed to adapt VLMs for classification \cite{gao2021clip,kim2021adapt,zhang2021tip,xing2023dual,wei2024benchmarking,wei2024snap}, object detection \cite{feng2022promptdet,gu2021open}, visual context parsing \cite{zhang2025cross}, disease diagnosis \cite{liu2025vision}, lightweight models \cite{cai2025prompt}.


\vspace{5pt}
\noindent \textbf{VLM Adaptation Methods.} 
We compare Prompt Logits Ensembling with existing adaptation methods as shown in Fig. \ref{performance}.
\textbf{Prompt-Tuning} \cite{dong2024cluster,han2024f,huang2024joint,lim2024dipex} focuses on only learning the prefix embeddings [$V_1$],[$V_2$]\ldots[$V_M$] of the prompts. The learned prefix embeddings can be combined with any class labels in the same context to form a text prompt ``[$V_1$],[$V_2$]\ldots[$V_M$] \{label\}”. 
While this approach offers flexibility, it suffers from a significant limitation: a single prompt often fails to capture the complexity and variance of the visual space, as demonstrated in Fig. \ref{pets_vis}. 
As an alternative to prompt-tuning, \textbf{Adapter-based methods} \cite{gao2024clip,zhang2024ta} learn to adjust the visual features and/or the text features with a residual connection. CLIP-Adapter \cite{gao2024clip} and Tip-Adapter \cite{zhang2022tip} are two representative adapter-based methods for CLIP.  Tip-Adapter builds a cache model that stores the visual features of training samples, which can also be seen as multiple prototypes for each class. The major problem is the limited visual samples in a few-shot learning setting. Very few training samples may reflect only a subset of the class distribution, resulting in a biased few-shot model. In comparison, prompts are initialized with diverse textual descriptions, providing more comprehensive coverage for class distribution. In addition, as CLIP-Adapter requires the visual samples to build prototypes, it is thus impossible to generalize to unseen classes. 
\textbf{Prompt Ensembling} \cite{radford2021learning,pratt2023does,wortsman2022robust} is the de facto method for zero-shot classification for CLIP. It averages the text features of various prompts associated with the class name to represent the classifier. The resulting class centroids will lose the local cluster information.
Our \textbf{Prompt Logits Ensembling} method offers distinct advantages over these approaches. By operating in the logits space rather than the feature space, it preserves the intrinsic cluster structure of class representations in CLIP. This design ensures that the nuanced class distributions are maintained, enhancing generalization and accuracy compared to the aforementioned methods.

\vspace{5pt}
{{\noindent \textbf{Relationship to Mixture of Experts (MoE)}. ~~Traditional MoE architectures involve a learnable gating network that dynamically routes each input to one or a few experts, adding stochastic behaviour and extra parameters to the system \cite{shazeer2017outrageously,fedus2022switch}. By contrast, CAPEL dispatches every image to all K prompt-specific sub-classifiers and then aggregates their logits through a lightweight, class-conditioned attention matrix. Through this design, we entirely remove instance-wise routing while still capturing latent class structure. Whereas MoE experts are usually full backbone blocks with large memory footprints, ``experts" of CAPEL are merely shallow linear heads (one per prompt) stacked on top of a frozen VLM, incurring minimum extra parameters. The training objectives also diverge. MoE models optimize routing-aware losses but do not explicitly enforce inter-expert semantic alignment. In comparison, CAPEL introduces a cluster-preserving conditional-entropy loss that sustains semantic diversity among prompts while aligning their logits with class-conditional clusters. Further, because CAPEL aggregates in logit space and learns prompt-level attentions, noisy prompts are automatically down-weighted, avoiding the catastrophic routing mistakes sometimes observed in MoE gating. Finally, the two paradigms are orthogonal and potentially complementary. An MoE backbone could be paired with CAPEL’s logit-ensemble and entropy regulariser for even greater scalability as an exciting avenue for future work.}}

\section{Preliminaries}
\noindent \textbf{Revisiting CLIP.}
Prompt engineering leverages the versatility of VLMs (CLIP in this work) to perform zero-shot classification. 
In essence, class names are combined with manually tailored textual templates to produce prompts. Consider $Y$ class names, represented as label$_y$, $y \in \{ 1,\ldots, Y\}$. Each class name is placed within a template, generating a prompt such as $\vh_y$ = ``a photo of a \{label$_y$\}". The CLIP text encoder processes this prompt $\texttt{TextEnc}(\cdot)$, and compute corresponding text features $\vw_y = \texttt{TextEnc}(\vh_y) \in \mathbb{R}^{D}$. Then, $\mW = [\vw_1^T,\ldots,\vw_Y^T]^T$ can be considered as the weights of a linear classifier. For any image $\mI$ to be classified, they are passed through the image encoder, $\texttt{ImageEnc}(\cdot)$, resulting in image features $\vx = \texttt{ImageEnc}(\mI) \in \mathbb{R}^{D}$. We can then predict the class of a test image with the prediction probability as:
\begin{equation}
\begin{gathered}
\tilde{y} = \text{argmax}_y P(y|\vx, \mW) = \frac{ \text{exp} (\tau\, \cos(\vx, \vw_y)  ) }{\sum^Y_{y=1} \text{exp} ( \tau\, \cos(\vx, \vw_y)   )},
\end{gathered}
\label{class_prob_CLIP}
\end{equation}
where $\tau$ is the temperature. During computing $\vw_y$, the class-specific training images are not required, thus enabling zero-shot recognition for any given class name.

\vspace{5pt}
\noindent \textbf{Prompt Ensembling.}
Considering that the training data of CLIP are image-text pairs, an image may match various sentences containing its class name. Zero-shot CLIP ViT-B/16 performance on ImageNet increases from 66.92\% to 68.57\% when using 80 hand-crafted prompts rather than ``A photo of \{\}" \cite{radford2021learning}. To use a set of hand-crafted prompts for zero-shot classification, the text features of the $K$ prompts composed with class names are averaged into a single vector to represent the class $y$ such that $\bar{\vw_y} = \frac{1}{K} \sum_{k=1}^{K} \vw_y^k$. To further improve these class prompts, Pratt \textit{et al.} \cite{pratt2023does} leveraged the power of large language models to create customized prompts that contain important discriminating characteristics of the image categories.  In conclusion, these methods leverage the power of prompt ensembling by averaging the diverse text features.

\begin{figure}[t]
    \centering
    \includegraphics[width=1.0\linewidth]{ 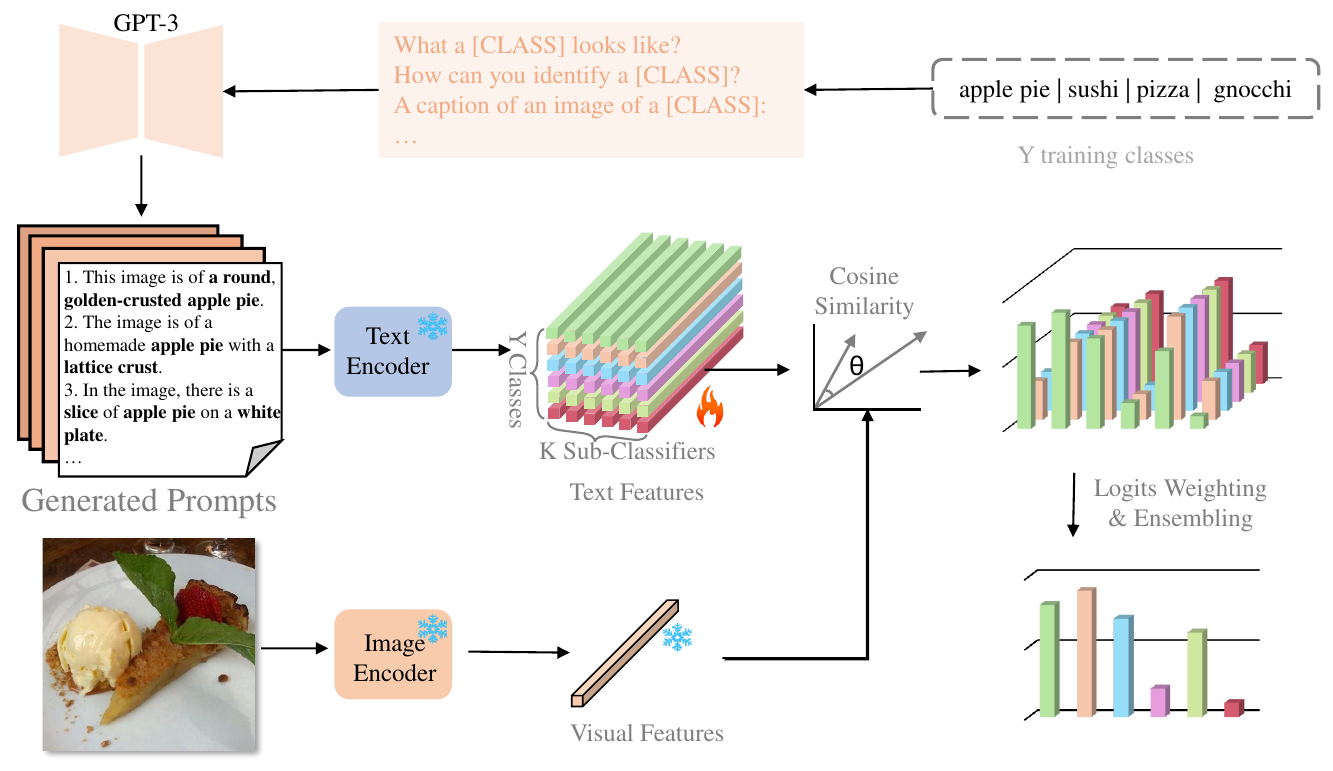}
    \caption{An illustration of Cluster-Aware Prompt Ensemble Learning for VLMs. We first leverage GPT-3 to generate K prompts that describe Y training classes. The prompts are extracted as text features to initialize the classifiers. We combine the visual features and the classifiers to get $Y \times K$ logits. Through \textbf{Cluster-Preserving Regularization} (Figure \ref{competition}) and \textbf{Adaptive Prompt Weighting}, we obtain the prediction for a particular class.}
    \label{fig:architecture}
\end{figure}

\section{Cluster-Aware Prompt Ensembling Learning}
In this section, we introduce Cluster-Aware Prompt Ensembling Learning (CAPEL). We first leverage GPT-3 to generate class-level descriptive prompts, which are then encoded into textual features using the CLIP text encoder. Next, we initialize an ensemble of classifiers with the textual features. Further, the classifiers are fine-tuned in a few-shot setting with cluster-preserving loss and an adaptive prompt weighting technique. 

\subsection{Classifier Initialization}
To acquire precise, articulate, and high-quality prompts to initialize the classifier weights, we leverage the pre-trained large language model GPT-3 \cite{brown2020language}. For each class in the datasets, GPT-3 generates an array of varied descriptions, ensuring broad coverage of potential visual interpretations for each category. 
Let $Y$ be the number of classes, $K$ the number of sub-prompts per class, and $D$ the text/image embedding dimension. For class $y\!\in\!\{1,\ldots,Y\}$
and prompt index $k\!\in\!\{1,\ldots,K\}$, we obtain the normalized text embedding $\vw_y^k$ from prompt $\vh_y^k$:
\begin{align}
\vw_y^k = \frac{\mathrm{TextEnc}(\vh_y^k)}{\bigl\|\mathrm{TextEnc}(\vh_y^k)\bigr\|_2} \in \mathbb{R}^{D}.
\end{align}
Stacking the $K$ sub-classifiers for class $y$ gives
\begin{align}
\mW_y  = \begin{bmatrix} (\vw_y^1)^\top 
\\ \vdots \\ 
(\vw_y^K)^\top \end{bmatrix} \in \mathbb{R}^{K\times D},
\mW = \mathrm{Stack}\!\left(\mW_1,\ldots,\mW_Y\right)
  \in \mathbb{R}^{Y\times K\times D}.
\end{align}
For matrix multiplication, we use the flattened view
$\mW^{\flat}\!\in\!\mathbb{R}^{(YK)\times D}$ obtained by row-wise concatenation.


\paragraph{Cost and reproducibility of LLM-generated prompts.}
To showcase the practical overhead of using large-language models (LLMs) to generate prompts, we provide the entire prompt bank, including 50 prompts per class for every dataset considered in this work through the \texttt{davinci-002} API. At the July-2025 pricing tier this required US\$7.8 in total. To ensure strict reproducibility, we release the JSON files of all prompts to our public repository. For scenarios in which external LLM access is impossible, in Table \ref{initialization} we show that starting from a small bank of handcrafted templates can also achieve relatively competitive performance. Finally, we acknowledge that future changes in API pricing or rate limits could hinder widespread adoption and therefore encourage further work on lightweight, on-device prompt generators.

\subsection{Prompt Logits Ensemble}
\label{prototype_ensemble}
Given a normalized image embedding $\vx\!\in\!\mathbb{R}^{D}$, the cosine similarity to each sub-classifier is defined elementwise by:
\begin{align}
\mZ_{y,k} = \tau\, \langle \vx, \vw_y^k \rangle
        = \tau\, \cos(\vx, \vw_y^k),
\qquad
\mZ \in \mathbb{R}^{Y\times K},
\end{align}
where $\tau$ is the temperature. 

We aim to make each sub-classifier of $y^{th}$ class, \textit{i.e.,} $\{\vw^1_y,\vw^1_y,\ldots,\vw^K_y\}$, specialize on distinct aspects of this class. In this way, given $K$ sub-classifiers, the averaged class logits over sub-classifiers  $\sum_{k=1}^K \mZ_{k} / K$ can be taken to compute the cross-entropy loss after Softmax layer like standard supervised learning.

\subsubsection{Adaptive Prompt Weighting}
The classifier weights are initialized by text features of the GPT-3 generated prompts, which inevitably involve ambiguous and flawed descriptors. To mitigate the impact of these flawed descriptors, our model is designed to assign lower confidence to such prompts while amplifying the influence of more accurate and representative ones.

To this end, we introduce an attention matrix to weight the varying significance among the diverse classifiers. Given $Y$ classes, each with $K$ prototypes, we formulate an attention matrix $\boldsymbol{\alpha}$. 
This matrix is employed to weigh the cosine similarities.
We consider the prediction logits from all sub-classifiers and take the weighted average logits to calculate the cross-entropy loss:
\begin{equation}
\begin{gathered}
  \ell_{avg} = - log \frac{ \text{exp} \Big( \sum_k \alpha_y^k \, \tau \, \cos\,(\vx, \vw_y^k)  \Big) }   {\sum_{y} \text{exp} \Big( \sum_k \alpha_y^k \, \tau \,\cos(\vx, \vw_y^k)  \Big) }. 
\end{gathered}
\label{avg_loss}
\end{equation}

\begin{figure}[t]
\begin{center}
\includegraphics[width=0.95\linewidth]{ 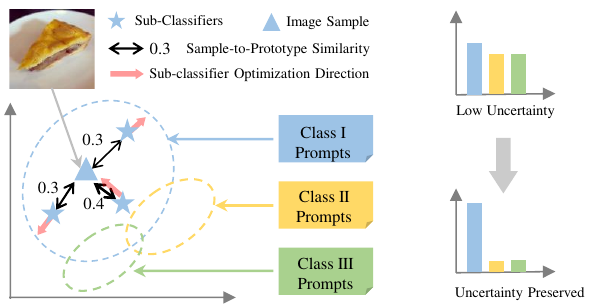}
\end{center}
\caption{Illustration of \textbf{Cluster-Preserving Regularization}. Each sub-prompt specializes in a distinct subset of the class, so that each image sample is supposed to be close to only one prompt, rather than close to all prototypes.}
\label{competition}
\end{figure}


\subsubsection{Cluster-Preserving Regularization with Conditional Entropy}
\begin{figure}[t]
\centering
\begin{subfigure}{0.495\linewidth}
\includegraphics[width=0.99\linewidth]{ 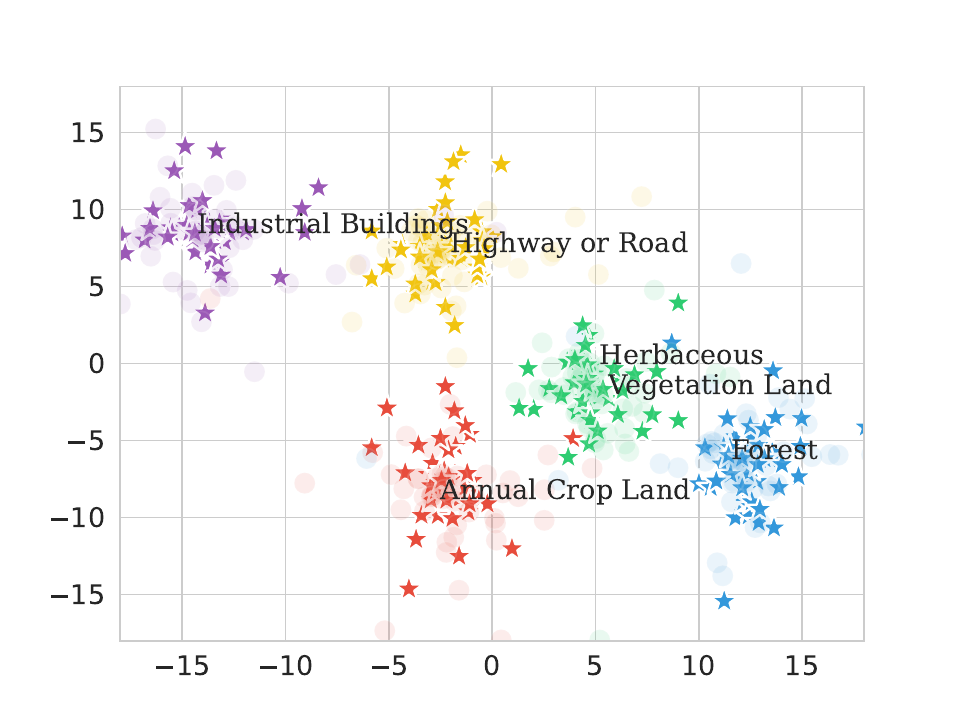}
\caption{Prototypes \textbf{with} Prototype Competition} 
\label{fig:vis-a}
\end{subfigure}
\hfill
\begin{subfigure}{0.495\linewidth}
\includegraphics[width=0.99\linewidth]{ 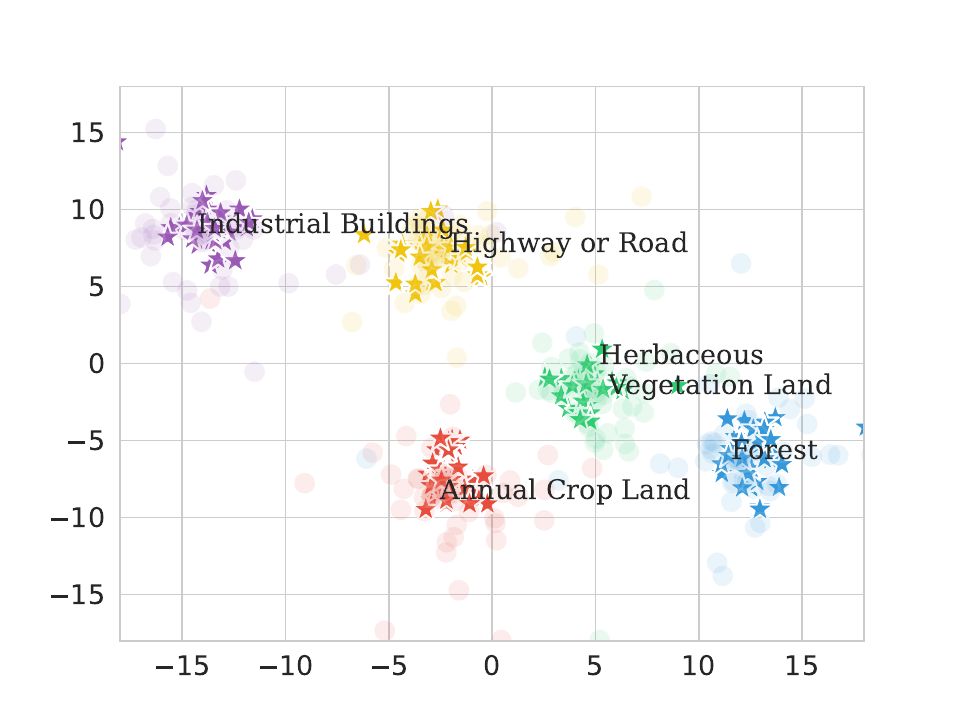}
\caption{ \textbf{without} Prototype Competition}
\label{fig:vis-b}
\end{subfigure}
\caption{t-SNE visualization of prototypes. Round dots are the initial prototypes, and asterisk dots are the learned prototypes. (a) With prototype competition, the learned prototypes are as diverse as original. (b) Without prototype competition, we observe that the prototypes tend to collapse into the class centroids. }
\label{fig:vis}
\end{figure}

The logits $\mZ$ produced by sub-classifiers are ultimately averaged for computing the cross-entropy loss as in Eq. \ref{avg_loss}, all the sub-classifiers receive the same feedback and will be optimized towards the same direction, as shown in Fig. \ref{fig:vis} (b). As a result, the class sub-classifiers tend to produce uniform logits. This phenomenon violates our motivation of making each sub-classifier of a class $y$ represent a cluster of the class. To mitigate this issue, as shown in Fig. \ref{competition}, we aim to make the closest sub-classifier (higher similarity) closer and other sub-classifiers far away. In terms of logits, we need to make the distribution of logits produced by sub-classifiers steeper. In this case, a training sample is \textbf{NOT} supposed to be close to every prompt. To do so, we resort to conditional entropy minimization to preserve the cluster nature. Let $k$ be the sub-classifier index. In essence, reducing the conditional entropy $H(k\,|\,\vx,y)$ allows $P(k\,|\,\vx,y)$ to be steeper:
\begin{align}
  H(k|\vx,y)& = - \mathbb{E}_{\vx,y} [\log P(k\,|\,\vx, y)] \\
            & = - \int \sum_{y} P(\vx, k\,|\,y ) \log P(k\,|\,\vx, y) dx.
\end{align}
Computing $H(k\,|\,\vx,y)$ requires modelling $P(\vx, k\,|y\,)$, but in supervised learning, we are motivated to model the conditional distribution. To avoid additional modeling by applying the plug-in principle, we replace the expectation for $(\vx,k|y)$ by the sample average. Thus, the empirical conditional entropy here can be written as:
\begin{align}
H_{emp}(k\,|\,\vx,y) = - \frac{1}{N} \sum_{i=1}^{N} \sum_{y} P \log P, \\ \text{where} \quad P = P(k \,|\,\vx_i, y),
\end{align}
where $N$ is the total number of training samples and the $\vx_i$ represents $i^{th}$ sample. Here, $P(k |\vx_i, y)$ corresponds to the logits $\mZ_y^k$ after Softmax. The objective function for prototype competition is designed by reducing the conditional entropy as follows:
\begin{align}
\ell_{pc} & = - \frac{1}{N}  \sum_{y} \frac{\text{exp}(\mZ_{y}^{k})}{\sum_{k} \text{exp}(\mZ_y^k)} \log \frac{\text{exp}(\mZ_{y}^{k})}{\sum_{k} \text{exp}(\mZ_y^k)}.
\end{align}

\begin{table*}[h!]
\centering
\caption{Numeric results of performance comparison between our proposed CAPEL and baseline methods in a few-shot learning setting}
\scalebox{0.85}{\begin{tabular}{clc|ccccccccccc}
\toprule
\rotatebox[origin=c]{45}{\#Shot } & \rotatebox[origin=c]{45}{Methods}& \rotatebox[origin=c]{45}{\textbf{Average}} & \rotatebox[origin=c]{45}{ImageNet} & \rotatebox[origin=c]{45}{UCF}  & \rotatebox[origin=c]{45}{Food101} & \rotatebox[origin=c]{45}{DTD} & \rotatebox[origin=c]{45}{StanfordCars} & \rotatebox[origin=c]{45}{Caltech} & \rotatebox[origin=c]{45}{Pets} & \rotatebox[origin=c]{45}{Flowers} & \rotatebox[origin=c]{45}{EuroSAT} & \rotatebox[origin=c]{45}{AirCraft} & \rotatebox[origin=c]{45}{SUN397}    \\
\midrule
\multirow{2}{*}{0} &CLIP & 65.31 & 66.73 & 67.67 & 85.86 & 44.03 & 65.64 & 93.31 & 89.13 & 70.73 & 48.11 & 24.72 & 62.56 \\
 &CuPL & 68.60 & 69.69 & 70.30 & 86.36 & 54.61 & 66.61 & 94.04 & 90.46 & 72.68 & 53.86 & 27.87 & 68.11 \\
\midrule
\multirow{9}{*}{1}     & CoOp & 66.57 & 66.32 &  71.68 & 81.26 & 47.04 & 64.39 & 90.87 & 86.10 & 82.14 & 54.85    & 25.95 & 61.64 \\
                       &ProDA&69.32 & 68.73 &  69.92 & 85.39 & 49.23 & 68.88 & 93.47 & 89.15 & 75.40 & 64.87     & 28.95 & 68.55 \\
                       &TIP-F&71.14 &  \underline{69.79} & 73.28 & 85.95 & 52.84 & 68.06 & 93.96 & 91.09 & 85.34 & 65.14 & 29.82 & 67.31 \\ 
                       &ProGrad&69.41 & 65.75 & 70.87 & 84.57 & 52.25 & 66.56 & 91.08 & 88.77 & \underline{85.55} & 66.19 & 26.97 & 64.99 \\
                       &MaPLe &69.69 & 64.89 & 74.49 & 83.70 & 51.30 & 67.17 & 93.85 & 91.10 & 78.56 &  \underline{68.74} & 28.74 & 64.08 \\
                       & TCP & 63.70 & 61.58 & 65.92 & 71.92 & 40.60 & 67.58 & 87.34 & 85.53 & 85.50 & 44.31 & 28.56 & 61.82 \\
                       & CuPL*      & 70.15 & 65.70 & \textbf{75.60} & 84.69 & 56.50 & 65.85 & 93.96 & 87.27 & \textbf{87.05} & 55.62 & 31.56 & 67.87\\
                       & CuPL*+WiSE & \underline{71.65} & 69.60 & 75.42 & \textbf{86.48} & \textbf{58.57} & \textbf{69.72} & \textbf{94.81} & 90.57 & 81.73 & 58.95 &  \underline{31.95} & \textbf{70.38}\\
                       \cmidrule(lr){2-14}
                        &  CAPEL& \textbf{72.97} & \textbf{70.48} &  \underline{75.57} &  \underline{86.26} &  \underline{57.92} &  \underline{68.92}  &  \underline{94.12} & \textbf{92.01} & {84.73} & \textbf{70.98} & \textbf{32.28} & \ \underline{69.39} \\
                        &  &  & $\pm2.7$ & $\pm3.3$ & $\pm2.4$ & $\pm3.8$ & $\pm1.7$ & $\pm1.4$ & $\pm3.9$ & $\pm1.1$ & $\pm0.4$ & $\pm2.1$ & $\pm0.9$ \\
\midrule
\multirow{9}{*}{2}     &CoOp & 70.23 & 67.87 &  74.33 & 82.32 & 49.88 & 69.28 & 93.51 & 86.26 & 89.77 & 61.14    & 31.29 & 66.86\\
                       &ProDA& 71.67 & 69.44 &  71.98 & 85.63 & 54.61 & 69.48 & 94.36 & 91.55 & 79.94 & 70.71     & 30.15 & 70.50\\
                       &TIP-F& 73.40 & 69.92 & 75.63 & 86.19 & 56.03 & 70.84 & 94.40 & 91.44 & 89.57 & 72.67 & 32.13 & 68.61 \\
                       &ProGrad& 71.77 & 67.05 & 73.75 & 83.79 & 54.49 & 71.62 & 93.92 & 90.92 & {89.97} & 66.44 & 29.91 & 67.65 \\
                       &MaPLe & 72.97 & 67.11 & 74.78 & 83.83 & 58.27 & 71.57 & 94.00 & \textbf{91.83} & 89.61 &  \underline{73.94} & 30.69 & 67.01\\
                       & TCP & 69.49 & 65.08 & 74.17 & 80.87 & 50.77 & 73.91 & 93.87 & 88.44 & \textbf{91.15} & 46.10 & 32.07 & 67.98\\
                       & CuPL*      & 72.68 & 65.67 &  \underline{78.03} & 84.50 & 60.46 &69.79 & 94.16 & 90.19 &  \underline{90.99} & 64.06 & 32.46 & 69.19\\
                       & CuPL*+WiSE &  \underline{74.13} &  \underline{70.16} & \textbf{78.35} &  \underline{86.55} & \textbf{61.82} & \textbf{72.42} & \textbf{95.09} &  \underline{91.82} & 85.87 & 68.84 &  \underline{33.27} &  \underline{71.28}\\
                       \cmidrule(lr){2-14}
                        & CAPEL& \textbf{74.90} & \textbf{70.74} & {77.93} & \textbf{86.55} &  \underline{59.99} &  \underline{72.34}  &  \underline{94.81} & 91.61 & 87.74 & \textbf{75.74} & \textbf{35.13} & \textbf{71.39}\\
                        &  &  & $\pm2.3$ & $\pm1.8$ & $\pm1.6$ & $\pm2.1$ & $\pm3.1$ & $\pm2.0$ & $\pm1.7$ & $\pm2.3$ & $\pm1.5$ & $\pm0.9$ & $\pm1.5$ \\
\midrule
\multirow{9}{*}{4}     &CoOp & 74.04 & 69.76 &  78.30 & 84.65 & 57.45 & 72.70 & 94.20 & 89.42 & 92.85 & 71.99     & 33.27 & 69.89\\
                       &ProDA& 74.27 & 70.96 &  76.05 & 86.38 & 61.23 & 69.99 & 95.01 & 92.23 & 86.24 & 73.87     & 32.67 & 72.38 \\
                       &TIP-F& 75.94 & 70.60 & 79.30 & 86.55 & 61.64 & 74.58 & 94.93 & 92.18 & 92.69 & 76.43 & 35.79 & 70.70 \\ 
                        &ProGrad& 74.95 & 69.40 & 77.05 & 85.97 & 58.10 & 75.20 & 94.56 & 92.07 & 92.61 & 74.46 & 33.58 & 71.47 \\
                        &MaPLe & 75.87 & 68.78 & 79.33 & 84.77 & 60.24 & 75.64 &  94.81 & 92.56 & 92.89 & 80.22 & 34.41 & 70.89\\
                        & TCP & 75.21 & 68.30 & 80.73 & 84.24 & 61.47 & \textbf{78.87} & 95.01 & 92.75 & \textbf{95.41} & 62.03 & 36.42 & 72.11 \\
                        & CuPL*      & 75.31 & 65.87 & \textbf{81.76} & 83.78 & 63.71 & 74.63 & 95.21 & 91.50 &  \underline{93.87} & 71.69 & 36.99 & 69.45 \\
                        & CuPL*+WiSE &  \underline{76.33} &  \underline{70.78} &  \underline{81.52} &  \underline{86.64} &  \underline{63.83} & 76.05 & \textbf{95.86} & \textbf{93.00} & 88.27 &  \underline{73.70} & \textbf{37.44} &  \underline{72.53}\\
                        \cmidrule(lr){2-14}
                         &  CAPEL& \textbf{77.64} & \textbf{71.56} & {81.23} & \textbf{86.95} & \textbf{64.60} &  \underline{76.26}   & \underline{95.62} &  \underline{92.61} & {92.89} & \textbf{81.56} &  \underline{37.35} & \textbf{73.44} \\
                        &  &  & $\pm1.4$ & $\pm1.7$ & $\pm0.8$ & $\pm0.9$ & $\pm1.4$ & $\pm1.6$ & $\pm3.1$ & $\pm0.5$ & $\pm4.1$ & $\pm2.0$ & $\pm1.2$ \\
\midrule
\multirow{9}{*}{8}     &CoOp & 76.55 & 70.81 &  79.12 & 85.14 & 64.42 & 76.42 & 95.25 & 92.01 & {95.93} & 76.46    & 34.15 & 72.33 \\
                       &ProDA& 76.76 & 71.89 &  79.75 &  \underline{87.07} & 67.08 & 71.88 & 95.05 & {93.16} & 92.33 & 77.81     & 34.32 &  73.99\\
                       &TIP-F&  \underline{78.80} & 71.93 & 82.18 & 86.65 & 67.14 & 77.85 & 95.33 & 92.50 & 95.29 & 82.07 & \underline{42.03} & 73.80 \\ 
                       &ProGrad& 76.86 & 70.90 & 80.68 & 86.76 & 60.22 & 78.53 & 95.21 & 92.23 & 93.79 & 75.86 & 38.16 & 73.12 \\
                       &MaPLe & 77.93 & 70.98 & 80.68 & 84.52 & 64.54 & 78.61 & 95.21 & 92.01 & 95.33 & \textbf{84.67} & 37.28 & 73.40  \\
                       & TCP & 78.39 & 70.98 &  \underline{83.43} & 85.84 & 68.09 & \textbf{81.59} & 95.46 &  \underline{93.13} &  \underline{96.59} & 70.35 & 41.88 & 74.92\\
                       & CuPL*      & 78.06 & 66.91 & 81.73 & 84.66 & 67.79 & 78.93 & 95.61 & 91.09 & \textbf{96.79} & 82.09 & 41.91 & 71.16\\
                       & CuPL*+WiSE & 78.62 &  \underline{72.17} & 82.05 & 86.99 &  \underline{68.09} & 79.68 & \textbf{96.35} & \textbf{93.21} & 91.23 & 80.67 & 40.24 &  \underline{74.23} \\
                       \cmidrule(lr){2-14}
                       & CAPEL& \textbf{80.14} & \textbf{72.53}  & \textbf{83.64} & \textbf{87.36} & \textbf{69.92} &  \underline{80.18}  &  \underline{95.98} & {93.08} & 95.66 &  \underline{84.57} & \textbf{43.32} &  \textbf{75.26} \\
                        &  &  & $\pm1.9$ & $\pm2.5$ & $\pm1.8$ & $\pm2.6$ & $\pm1.5$ & $\pm3.1$ & $\pm1.8$ & $\pm2.4$ & $\pm1.6$ & $\pm0.9$ & $\pm3.0$ \\
\midrule
\multirow{9}{*}{16}    &CoOp &78.82 & 71.92 &  81.95 & 86.32 & 69.42 & 78.70 & 95.42 & 92.80 & 95.94 & 84.44  & 35.89 & 74.28  \\
                       &ProDA&79.19 & 72.93     &  82.69 & 87.24 & 70.09 & 75.09 & 95.09 &  \underline{93.54} & 96.02 & 85.64    & 37.17 & 75.64 \\
                       &TIP-F&81.11 &  \underline{73.26} & 83.58 & 87.27 &  \underline{71.75} & 82.86 & 95.90 & 93.16 & 96.31 &  \underline{87.60} & 44.13 & 76.39 \\ 
                       &ProGrad &79.64 & 72.31 & 82.98 & 87.16 & 66.73 & 81.32 &  \underline{96.27 }& 93.00 & 95.37 & 84.25 & 41.88 & 74.79 \\
                       &MaPLe &80.94 & 72.24 & 83.74 & 85.58 & 70.74 & 83.22 & 96.19 & 93.23 & 96.91 & 87.56 & 45.33 &75.62 \\
                       &TCP &  \underline{81.29} &73.14 &  \underline{85.30}& 87.03 &71.69& \textbf{85.50} &96.15&92.91&97.44&83.19&45.07&  \underline{76.77}  \\
                       & CuPL*      & 79.59 & 67.75 & 84.59 & 84.36 & 70.57 & 82.81 & 95.78 & 91.33 & 93.02 & 85.13 &  \underline{46.33} & 73.79 \\
                       & CuPL*+WiSE & 80.90 & 72.91 & 84.80 &  \underline{87.29} & 70.80 & 82.86 & 96.19 & 92.61 &  \underline{97.69} & 83.85 & 44.79 & 76.11 \\
                       \cmidrule(lr){2-14}
                       &  CAPEL & \textbf{82.45} & \textbf{73.73} & \textbf{85.51} & \textbf{87.81} & \textbf{73.46} &  \underline{84.40}  & \textbf{96.55} & \textbf{93.57} & \textbf{97.97} & \textbf{88.73} &  \textbf{48.09} & \textbf{77.11} \\
                        &  &  & $\pm1.1$ & $\pm2.1$ & $\pm0.8$ & $\pm1.3$ & $\pm1.8$ & $\pm2.1$ & $\pm0.8$ & $\pm1.6$ & $\pm0.8$ & $\pm1.0$ & $\pm1.1$ \\
\bottomrule
\end{tabular}}
\label{numericfewshot}
\end{table*}

\subsubsection{Training and Inference}
The trainable parameters in our method include the sub-classifiers $ \mW \in \mathbb{R}^{Y \times K \times D}$, along with the attention matrix $\boldsymbol{\alpha} \in \mathbb{R}^{Y \times K}$. The overall training objective is:
\begin{equation}
\begin{gathered}
  \ell_{overall} = \ell_{ce} + \lambda\ell_{pc},
\end{gathered}
\label{overall_loss}
\end{equation}
where $\lambda$ is the loss weight. 
At test time, the class distribution of a test image can be computed as follows:
\begin{equation}
\begin{gathered}
    P(y|\vx) = \frac{ \text{exp} \Big( \sum_k \alpha_y^k \, \tau \, \cos(\vx, \vw_y^k) \Big) }
  {\sum_{y} \text{exp} \Big( \sum_k \alpha_y^k \, \tau \,\cos(\vx, \vw_y^k) /  \Big) },  
\end{gathered}
\label{class_prob_prototypes}
\end{equation}
where we consider the weighted average cosine similarities of the prototypes of every class. The class label of a test image $\vx$ is then deduced as $\tilde{y} = \argmax_{y}P(y\,|\,\vx)$.

\begin{table*}[t]
  \centering
  \caption{{{Efficiency comparison on a single NVIDIA RTX 3090 (\textbf{batch}=64, \textbf{shot}=16). \textbf{\textit{Pruned}} keeps heads with top 10 prompts with the highest weights.  }}}
  \label{tab:efficiency}
\scalebox{0.9}{ {{ \begin{tabular}{@{}clcccccc@{}}
    \toprule
      \multirow{1}{*}{\textbf{Dataset}}
      & \multirow{1}{*}{\textbf{Model}} 
      & \multirow{1}{*}{\textbf{\# Prompts}} 
      & \multirow{1}{*}{\textbf{Acc.\,(\%)}} 
      & \textbf{GFLOPS} & \textbf{Model Size} & \textbf{Train Cost} \\ 
   (classes) & (ViT-B/16, $224^{2}$) & & & (32-bit) 
      & (MB) &(GPU-mins) \\
    \midrule
    \multirow{5}{*}{\rotatebox[origin=c]{0}{DTD}} & CoOp & 1 & 69.42 & 2928.71 & 308.91 & 2.64  \\
   \multirow{5}{*}{\rotatebox[origin=c]{0}{(47)}} & TCP & 1 & 71.69 &  153.74 & 241.06 & 2.76  \\
    & CuPL* + WiSE  & 1   & 70.80    & \textbf{16.87} & \textbf{237.67} & 1.79 \\
    & \textbf{CAPEL (full)}   & 50  & \textbf{73.46}  & \textbf{16.87}($+$0\%) & 239.92 ($+$0.9\%) & 1.80 ($+$0.01) \\
    & \textbf{CAPEL (pruned)} & $10$& 73.31 ($-$0.15) & \textbf{16.87}($+$0\%) & 238.08 ($+$0.2\%) & - \\ 
    \midrule
    \multirow{5}{*}{\rotatebox[origin=c]{0}{ImageNet}} & CoOp & 1 & 71.92 & 2928.71 & 308.91 & 428.40\\
    \multirow{5}{*}{\rotatebox[origin=c]{0}{(1000)}} & TCP & 1 & 73.14 & 2929.04 & 297.83 & 363.6\\
    & CuPL* + WiSE  & 1        & 72.91               & \textbf{16.87} & \textbf{238.60} & 25.30 \\
    & \textbf{CAPEL (full)}   & 50       & \textbf{73.73}      & 16.89 ($+$0.1\%) & 286.54 ($+$16.7\%) &  27.80 ($+$2.5) \\
    & \textbf{CAPEL (pruned)} & $10$& 73.41 ($-$0.32)     & 16.87 ($+$0\%) & 247.40 ($+$3.5\%) & - \\
    \bottomrule
  \end{tabular}}}}
  \label{table:1}
\end{table*}

\begin{table*}[t]
    \centering
    \begin{minipage}{0.45\textwidth} 
        \centering
        \caption{Comparison of CAPEL and baselines on domain generalization from the 16-shot learning setting. }
\resizebox{0.99\textwidth}{!}{\begin{tabular}{ l c  cccc  c}
\toprule 
  & Source  &  \multicolumn{4}{c}{\textit{Target}} & {}  \\ 
\cmidrule(lr){2-2} \cmidrule(lr){3-6} \cmidrule(lr){7-7} 
& {IN}  & {-V2} & {-S} & {-A} & {-R} & {AVG}\\
\hline
CLIP    & 66.73 &  60.83  &  46.15  & 47.77 & 73.96 & 59.09\\
CoOp    & 71.51 &  64.20  &  47.99  & 49.71 & 75.21 & 61.72\\
ProDA   & 72.93 &  \underline{64.94}  &  48.90  & 50.23 & 75.81 & 62.56  \\ 
TIP-F   & \underline{73.26} &  64.82  &  49.05  & 48.97 & 75.82  & 62.38\\
ProGrad & 72.31 &  60.03  &  45.32  & 50.32 & 76.23  & 60.84\\
MaPLe   & 70.72 &  64.07  &  {49.15}  & \underline{50.90} & \underline{76.98} & 62.36\\
TCP & 71.20 & 64.60 & \textbf{49.50} & \textbf{51.20} & 76.73 & \underline{62.65}\\
CuPL* & 67.75 & 62.13 & 47.08 & 48.35 & 74.65 & 59.99\\
CuPL*+W & 72.91 & 64.90 & 49.08 & 49.88 & 76.23 & 62.60\\
\midrule
\rowcolor{tabhighlight} CAPEL    & \textbf{73.73} & \textbf{65.22} & \underline{49.38} & {50.80} & \textbf{77.09} & \textbf{63.24}\\
\bottomrule
\end{tabular}}
        \label{tab:left_table}
    \end{minipage}%
    \hfill
    \begin{minipage}{0.5\textwidth} 
        \centering
        \caption{Ablation study of CAPEL on ImageNet, DTD, and EuroSAT. Results showcase the importance of each component.}
\resizebox{0.99\textwidth}{!}{\begin{tabular}{ c|c|c|c | ccc }
\toprule 
\multicolumn{4}{c}{{Components}}  & \multicolumn{3}{c}{{Datasets}}  \\
\cmidrule(lr){1-4} \cmidrule(lr){5-7} 
\multirow{2}{*}{\rotatebox[origin=c]{60}{\shortstack{Fine-\\Tuning}}}     
& \multirow{2}{*}{\rotatebox[origin=c]{60}{\shortstack{Logits\\Ensemble}}}           
& \multirow{2}{*}{\rotatebox[origin=c]{60}{\shortstack{Cluster-\\Preserving}}}            
& \multirow{2}{*}{\rotatebox[origin=c]{60}{\shortstack{Prompt-\\Weighting}}}         
& \multirow{2}{*}{\rotatebox[origin=c]{60}{\shortstack{ImageNet}}} 
& \multirow{2}{*}{\rotatebox[origin=c]{60}{\shortstack{DTD}}} 
& \multirow{2}{*}{\rotatebox[origin=c]{60}{\shortstack{EuroSAT}}} \\
&   &    &     &                            &                      &  \\
&   &    &     &                            &                      &  \\
&   &    &     &                            &                      &  \\
\cmidrule(lr){1-4} \cmidrule(lr){5-7} 
 & & & & 
69.69 & 54.61 & 53.86  \\
&\checkmark&  & &
69.95 & 56.28 & 55.10 \\
\checkmark&&&  &
67.75 & 70.57 & 85.13 \\
\checkmark&\checkmark&&& 
72.02 & 71.23 & 85.33 \\ 
\checkmark& \checkmark& \checkmark & & 
\underline{73.10} & \underline{72.82} & \underline{86.23} \\
\checkmark& \checkmark&  & \checkmark& 
72.16 & 72.12 & 85.66 \\    
\midrule
\rowcolor{tabhighlight}  \checkmark&\checkmark& \checkmark & \checkmark & 
\textbf{73.73} & \textbf{73.46} & \textbf{88.73} \\
\bottomrule 
\end{tabular}}
        \label{tab:right_table}
    \end{minipage}
\end{table*}

\section{Experiments}
\subsection{Settings}
Our experiments explore visual models with language as internal representations. We show results for Cluster-Aware Prompt Ensembling Learning (CAPEL) as class representations for object recognition. 
We compare our methods against prompt tuning methods (CoOp \cite{zhou2022learning}, ProDA \cite{lu2022prompt},  MaPLe \cite{khattak2023maple}, ProGrad \cite{zhu2023prompt}, TCP \cite{yao2024tcp}), adapter-based method TIP-Adapter-F \cite{zhang2022tip}, prompt ensembling methods (CuPL \cite{pratt2023does} and WiSE \cite{wortsman2022robust}). 
CuPL* represents the fine-tuning version of the original zero-shot method CuPL. CuPL*+WiSE represents that we interpolate the fine-tuned and zero-shot models as indicated in WiSE. Both CuPL* CuPL*+WiSE are trained with 100 epoches, which is same as our proposed CAPEL.
These are the most recently established methods for adapting CLIP to downstream object recognition tasks. We consider a variety of image domains, including a scene recognition dataset SUN397 \cite{xiao2010sun}; an action recognition dataset UCF101 \cite{soomro2012ucf101}; a satellite image dataset EuroSAT \cite{helber2019eurosat}; a texture dataset DTD \cite{cimpoi2014describing}; two coarse-grained object datasets, ImageNet \cite{deng2009imagenet} and Caltech101 \cite{fei2004learning} and five fine-grained datasets, OxfordPets \cite{parkhi2012cats}, StanfordCars \cite{krause20133d}, Flowers102 \cite{nilsback2008automated}, Food101 \cite{bossard2014food}, and FGVCAircraft \cite{maji2013fine}. For domain generalization, we choose ImageNet as source domain dataset and evaluate performance on four target datasets: ImageNetV2 \cite{recht2019imagenet}, ImageNet-Sketch \cite{wang2019learning}, ImageNet-A \cite{hendrycks2021natural}, ImageNet-R \cite{hendrycks2021many}.
We use a few-shot training strategy in all experiments at 16 shots which are randomly sampled for each class. We apply CAPEL on a pre-trained ViT-B/16 CLIP unless stated otherwise. We generate 50 prompts per category for all 11 datasets with the GPT-3 model \cite{brown2020language}. The textual features of prompts are fine-tuned for 100 epochs. The batch size is set to 64. We use an SGD optimizer with a learning rate of 2e-3. All experiments are conducted on an NVIDIA A6000 GPU. We report the performance over 3 runs of different random seeds. We mainly instruct GPT-3 with the five prompts to  generate the prototypes. {y} is the category name and {length} is the expected sentence length of the generated prompts.

\noindent 1. ``Describe a photo of \{y\} in one sentence, no more than \{length\} words."

\noindent2. ``How does a \{y\} look like? Answer in no more than \{length\} words."

\noindent3. ``Summarize visual features of \{y\} in no more than \{length\} words."

\noindent4. ``Tell me what \{y\} looks like in a short sentence, less than \{length\} words."

\noindent5. ``Use less than \{length\} words to outline the look of \{y\}."



\subsection{Few-Shot Learning Results}
We conduct experiments on all classes of the 11 datasets and compare our method with CoOp, ProDA, TIP-Adapter-F, and ProGrad, MaPPLe, TCP, CuPL, CUPL+WiSE in a few-shot learning setting. The results are reported in Table \ref{numericfewshot}. CAPEL respectively gained 1.32\%, 0.77\%, 1.31\%, 1.34\%, and 1.16\% performance boost over the second-best methods at 1, 2, 4, 8, and 16 shots. The performance improvement verified the effectiveness of prompt logits ensembling over prompt feature ensembling. The ensembling method CuPL*+WiSE achieves the second-best performance in 1, 2, 4 shots settings, while Adapter-based method TIP-F achieves the second-best performance with 8 shots. The latest prompt tuning method TCP achieves the second-best performance with 16 shots. Our method CAPEL performance the best across all few-shot settings, demonstrating its robustness to different shots.

\begin{table*}[t]
\setlength{\tabcolsep}{3pt}
\caption {Performance comparison between different prompt initialization methods.}
\vspace{-10pt}
\begin{center}
\scalebox{0.99}{\begin{tabular}{c|c|ccccccccccc}
\toprule
Weight Init  & Average & ImageNet & Stan/Cars & SUN397 & Cal/101 & Oxf/Pets & Oxf/Flo & DTD & FGVC & EuroSAT & UCF101 & Food101 \\
\midrule
Kaiming   & 75.17 & 54.71 & 78.27 & 70.92 & 93.75 & 75.22 & 96.71 & 65.28 & 45.90 & 86.40 & 79.94 & 79.77 \\
Xavier    & 76.62 & 66.67 & 79.14 & 71.13 & 93.63 & 75.91 & 96.59 & 65.96 & 47.34 & 86.04 & 81.07 & 79.37 \\
Template  & 80.30 & 71.53 & 82.45 & 72.81 & 95.15 & 92.08 & 96.48 & 70.22 & 45.71 & 86.78 & 83.67 & 86.44 \\
\midrule
\rowcolor{tabhighlight} GPT-3     & \textbf{82.48} & \textbf{73.73} & \textbf{84.80} & \textbf{77.11} & \textbf{96.55} & \textbf{93.57} & \textbf{97.97} & \textbf{73.46} & \textbf{48.09} & \textbf{88.73} & \textbf{85.51} & \textbf{87.81} \\
\bottomrule 
\end{tabular}}
\end{center}
\label{initialization}
\vspace{-10pt}
\end{table*}
\subsection{Comparison on Computational Efficiency}
Table \ref{table:1} compares our CAPEL to three recent prompt-based baselines on DTD (47 classes) and ImageNet (1,000 classes). CoOp and TCP rely on a single prompt, while CuPL* + WiSE retains one prompt per class but averages features. CAPEL instead instantiates 50 sub-prompts and ensembles them in logit space. Despite the larger prompt set, on DTD, CAPEL matches the low compute footprint of CuPL ($\sim$\! 16.87 GFLOPS) and keeps the parameter overhead below one per cent. It raises 16-shot accuracy to 73.46\%, a 1.77\% gain over the strongest baseline (TCP), and needs only 1.80 GPU-minutes for training, almost identical to CuPL. A simple pruning step that discards heads with low attention weights trims the active prompts to ten, shaving 1.84 \% of parameters while sacrificing just 0.15 points of accuracy and leaving compute unchanged. The same pattern appears on ImageNet: CAPEL improves top-1 accuracy from 73.14 \% (TCP) to 73.73 \%, yet still operates at $\sim$16.89 GFLOPS, which is two orders of magnitude lower than CoOp or TCP, and adds only 2.5 GPU-minutes of training time. After pruning, the model size is just 3.5 \% above CuPL while the accuracy drop remains within 0.32 points. Overall, CAPEL delivers the best accuracy–efficiency trade-off among all methods considered.

\subsection{Domain Generalization}
The effectiveness of CAPEL in generalizing to out-of-distribution datasets is demonstrated when compared with the baseline methods. We evaluate the direct generalization ability of a model trained on ImageNet to several out-of-domain datasets. The results, presented in Table \ref{tab:left_table}, show that CAPEL outperforms existing methods on both source domain and target domains, except on ImageNet-A and ImageNet-S. The results confirm that CAPEL is also domain-generalizable.

\subsection{Ablation Study} 
To dissect the impact of each component within the proposed CAPEL framework, we conduct an ablation study. In Table \ref{tab:right_table}, we deconstruct the full framework into its constituent parts:  fine-tuning (fine-tuning the textual features of these class sub-prompts), logits ensembling (employing 50 distinct class sub-prompts per class), adaptive prompt weighting (leveraging an attention matrix to adjust the weighting of different sub-classifiers), and cluster-preserving regularization (leveraging conditional entropy to penalize a uniform distribution of logits). The effectiveness of each component was evaluated with the 16-shot setting on from ImageNet, DTD, and EuroSAT datasets. The results underscore the distinct and vital role each component plays in our comprehensive framework. In comparison to the baseline CuPL that average the 50 prompts to represent a class, our comprehensive approach manifests significant performance enhancements across the three datasets, recording improvements of 4.04\%, 18.85\%, and 34.87\%, respectively. 
Notably, logits ensembling results in substantial performance uplifts of 0.26\%, 1.67\%, and 1.24\%. 
After fine-tuning, logits ensembling achieves further 2.07\%, 0.66\%, and 0.2\% improvement. 
By implementing an adaptive attention matrix to weight the importance of each sub-prototype within a class, we further augmented our performance by increments of 0.14\%, 0.89\%, and 0.33\%. The integration of our cluster-preserving loss results in additional accuracy increments of 1.08\%, 0.64\%, and 2.50\%.

\vspace{5pt}
\noindent \textbf{Impact of Sub-Classifier Initialization}
To demonstrate the effectiveness of the GPT-3 generated prompts, we apply different weight initialization methods for classifiers in a few-shot learning setting. Specifically, we compare with Xavier \cite{glorot2010understanding}, Kaiming \cite{he2015delving}, and template prompts. For template prompts, we extract the text features of ``\texttt{A photo of \{\}}” as the same weights for all 50 class sub-classifiers. The performance comparison is reported in Table \ref{initialization}. On average, GPT-3 generated prompts respectively gain 7.31\%, 5.86\%, and  2.18\% performance boost over Kaiming, Xavier, and template prompts. Note that all three compared methods use 50 prompts for fair comparison.

\begin{table}[t]
\setlength{\tabcolsep}{1.2pt}
\renewcommand{\arraystretch}{0.61}
\centering
\caption{Effect of prompt-generator choice on CAPEL accuracy (top-1 \%). 
Prompts were produced by five LLMs and used without any other changes. 
Results span three datasets and 1/4/16-shot settings. Bold indicates the best value in each row. 
Average accuracies across all nine settings differ by at most 0.43\%, showing that CAPEL is insensitive to the specific LLM used to draft prompts.
}
\scalebox{0.8}{
{{\begin{tabular}{lcccccc}
\toprule
    Datasets & Shots & GPT-3 & GPT-4  &  Llama-4-Scout & Qwen-Turbo & Gemini-2.5-Pro  \\
\midrule
\multirow{3}{*}{{DTD}}      
        & 1     & \textbf{57.92} & 57.42 & 57.65 & 57.55 & 57.49\\ 
        & 4     & 64.60         & 64.12 & \textbf{65.14} & 63.87 & 64.57 \\
        & 16    & 73.46         & 73.75 & \textbf{73.92} & 73.81 & 72.91\\
       \hline
\multirow{3}{*}{OxformPets} 
        & 1     & 92.01 & 91.74 & 90.87 & 91.28 & \textbf{93.78}  \\
        & 4     & \textbf{92.61} & 92.17 & 92.21 & 92.37 & 91.48\\
        & 16    & \textbf{93.57} & 92.98 & 93.48 & 92.48 & 93.34\\
        \hline
\multirow{3}{*}{SUN397}  
        & 1     & \textbf{69.39} & 68.87 & 68.99 & 68.91 & 67.89  \\
        & 4     & \textbf{73.44} & 73.18 & 73.12 & 72.80 & 72.14 \\
        & 16    & 77.11 & 76.82 & 76.15 & 77.08 & \textbf{77.28} \\
        \hline
        Avg & & \textbf{77.12} & 76.78 & 76.84 & 76.69 & 76.76\\
\bottomrule
\end{tabular}}}}
\label{VLMs1}
\vspace{-10pt}
\end{table}

\begin{table*}[t]
\centering
\caption{Generalisability across CLIP backbones (ResNet and ViT). CAPEL achieves the best average accuracy over eight variants, including RN50/RN101.}
\vspace{-5pt}
\scalebox{0.80}{
\begin{tabular}{l|c|cccccccc}
\toprule
    Models & \rotatebox[origin=c]{40}{Average} &  \rotatebox[origin=c]{40}{{{RN50}}}&    \rotatebox[origin=c]{40}{{{RN101}}}&\rotatebox[origin=c]{40}{ViT-B/32} & \rotatebox[origin=c]{40}{ViT-B/16} & \rotatebox[origin=c]{40}{ViT-L/14}  & \rotatebox[origin=c]{40}{ViT-L/14@336} & \rotatebox[origin=c]{40}{LAION ViT-B-32}  & \rotatebox[origin=c]{40}{BLIP ViT-B} \\
\midrule
Zero-Shot  & 65.83 & 60.33 & 62.53 & 63.80 & 68.73 & 75.30 & 76.20 & 66.60 & 53.15\\
CoOp       & 68.76 & 62.95 & 66.60 & 66.85 & 71.92 & 78.21 & 76.23 & 71.46 & 55.87 \\
ProDA      & 70.75 & 64.28 & 67.63 & 67.79 & 72.93 & 78.85 & 78.32 & 72.98 & 63.21 \\
MaPLe      & 71.27 & 65.21 & 68.32 & 68.21 & 72.24 & 78.96 & 79.48 & 72.85 & 64.89 \\
TIP-F      & 71.66 & 65.51 & 68.56 & 68.65 & 73.26 & 79.14 & 79.82 & 73.18 & 65.19 \\
TCP        & 71.63 & 65.04 & 68.19 & 67.88 & 73.14 & 79.06 & 80.30 & 73.16 & 66.27 \\
CuPL*      & 65.3 & 60.08 & 62.32 & 62.96 & 67.75 & 75.22 & 75.64 & 64.78 & 53.65 \\
CuPL*+WiSE & 71.48 & 65.18 & 67.72 & 67.98 & 72.91 & 78.99 & 79.70 & 73.45 & 65.87 \\
\midrule
 CAPEL      & \textbf{72.34} & \textbf{66.07} & \textbf{68.93} & \textbf{69.03} & \textbf{73.73} & \textbf{79.63} & \textbf{80.44} & \textbf{74.12} & \textbf{66.80} \\
\bottomrule
\end{tabular}}
\label{Variants}
\end{table*}

\begin{table*}[t!]
\centering
\caption{{{Few-shot anomaly detection on \textsc{MVTec} and \textsc{VisA} (top-1 \%). 
Values are mean $\pm$ std over runs.}}}
\label{tab:mvtec-visa-fewshot}
{\scalebox{0.99}{
\begin{tabular}{@{} l l ccc ccc @{}}
\toprule
\multirow{2}{*}{Method} & \multirow{2}{*}{Venue} 
& \multicolumn{3}{c}{MVTec} & \multicolumn{3}{c}{VisA} \\
\cmidrule(lr){3-5}\cmidrule(lr){6-8}
& & 1-shot & 2-shot & 4-shot & 1-shot & 2-shot & 4-shot \\
\midrule
SPADE~\cite{cohen2020sub}          & arXiv'2020 & $81.0\pm2.0$ & $82.9\pm2.6$ & $84.8\pm2.5$ & $79.5\pm4.0$ & $80.7\pm5.6$ & $81.7\pm3.4$ \\
PaDiM~\cite{defard2021padim}          & ICPR'2020  & $76.6\pm3.1$ & $78.9\pm3.1$ & $80.4\pm2.4$ & $62.8\pm5.4$ & $67.4\pm5.1$ & $72.8\pm2.9$ \\
PatchCore~\cite{roth2022towards}  & CVPR'2022  & $83.4\pm3.0$ & $86.3\pm3.3$ & $88.2\pm2.6$ & $79.9\pm2.9$ & $81.6\pm4.0$ & $85.3\pm2.1$ \\
WinCLIP$^{++}$~\cite{jeong2023winclip} & CVPR'2023  & $93.1\pm2.0$ & $94.4\pm1.1$ & $95.2\pm1.3$ & $83.8\pm4.0$ & $84.6\pm2.4$ & $87.3\pm1.8$ \\
RWDA~\cite{tamura2023random}            & BMVC'2023  & $93.3\pm0.5$ & $94.0\pm0.7$ & $94.5\pm0.7$ & $83.4\pm1.7$ & $85.6\pm1.4$ & $86.6\pm0.9$ \\
FastRecon~\cite{fang2023fastrecon}  & ICCV'2023  & --           & $91.0$       & $94.2$       & --           & --           & --          \\
PromptAD~\cite{li2024promptad}          & CVPR'2024         & $94.6\pm1.7$ & $95.7\pm1.5$ & $96.6\pm0.9$ & $86.9\pm2.3$ & $88.3\pm2.0$ & $89.1\pm1.7$ \\
\midrule
CAPEL(ours)          &  PR'2025        & \textbf{94.9$\pm$1.5} & \textbf{96.8$\pm$2.1} & \textbf{97.7$\pm$1.2} & \textbf{86.2$\pm$1.7} & \textbf{88.7$\pm$0.8} & \textbf{89.8$\pm$0.5} \\
\bottomrule
\end{tabular}}}
\vspace{2pt}
\label{industy}
\end{table*}

\vspace{10pt}
\noindent \textbf{Impact of Prompts Generated by Different LLMs.} ~~ We follow the same prompts as shown in Section 5.1 in the manuscript to instruct LLMs for generating class descriptions, including GPT-4 \cite{achiam2023gpt}. Llama-4-Scout \cite{meta2025llama4scout}, Qwen-Turbo \cite{team2024qwen2}, Gemini-2.5-Pro \cite{deepmind_gemini25pro_2025}. By replacing the existing class descriptions generated by GPT-3, we have obtained performance results as shown in Table \ref{VLMs1}. It shows that CAPEL is remarkably insensitive to which large-language model is used to draft the initial text prompts. Across three datasets, including DTD, Oxford Pets and SUN397, and three few-shot settings (1, 4, 16), the span between the best and worst average accuracies among the five LLMs is just 0.43 \% (77.12 \% for GPT-3 vs. 76.69 \% for Qwen-Turbo). Per-dataset trends are likewise narrow. on DTD, the most advanced model Llama-4-Scout surpasses GPT-3 by only 0.46\% in the 16-shot setting, while on Oxford Pets and SUN397 the gains occasionally swing the other way. For example, Gemini-2.5-Pro tops the 1-shot Oxford Pets run by 1.77\%, but falls behind in the 4 and 16-shot cases. These fluctuations are well within one standard deviation of CAPEL’s inherent run-to-run variance, indicating that prompt diversity, once fed into our logit-space ensemble and adaptive weighting, dominates any marginal quality differences among modern LLMs. In short, CAPEL delivers essentially the same performance whether its prompts originate from GPT-3, GPT-4, or state-of-the-art open-source alternatives. More powerful LLMs do not translate into systematically higher accuracy, which shows the robustness of our method to the prompt sources.

\noindent \textbf{Generalizability Across VLM Variants}
We conducted extensive experiments with various CLIP models and other VLMs, as detailed in Table \ref{Variants}. Our method's consistent performance improvement across different architectures, including ViT-B/16, ViT-B/32, ViT-L/14, ViT-L/14@336, LAION's CLIP replication \cite{schuhmann2022laion}, and the BLIP model \cite{li2022blip}, attests to its robustness. These results underscore our method's adaptability to different model architectures and training datasets. This is particularly significant given that each of these variants and models has unique characteristics and was trained on diverse datasets.

\begin{table}[t]
\centering
\setlength{\tabcolsep}{3pt}
\caption{{{Medical-domain results on \textsc{Kvasir} and \textsc{KneeXray} across 1/4/16 shots. CAPEL outperforms CLIP adaptation baselines, including BiomedCoOp.}}}
\scalebox{0.8}{
\begin{tabular}{lccccccc}
\toprule
    Datasets & Shots & CLIP & CoOp  & TIP-F&  ProGrad & BiomedCoOp & \textbf{CAPEL}  \\
\midrule
\multirow{5}{*}{Kvasir}      
        & 1     & \multirow{5}{*}{54.58} & 58.2 & 59.19 & 60.78 & 62.17 & \textbf{62.78}\\ 
        & 2     &  & 64.86 & 64.22 & 64.70 & \textbf{67.25} & 66.98\\
        & 4     &  & 70.78 & 69.94 & 70.00 & 74.08 & \textbf{74.89}\\
        & 8     &  & 77.14 & 75.86 & 76.03 & 77.72 & \textbf{78.18}\\
        & 16    &  & 77.88 & 78.00 & 75.88 & 78.89 & \textbf{79.99}\\
       \hline
\multirow{5}{*}{KneeXray} 
        & 1     &  \multirow{5}{*}{29.53} & 24.96 & 30.01 & 30.09 & 36.13 & \textbf{37.74}\\
        & 2     & & 25.89 & 28.38 & 23.83 & 37.72 & \textbf{38.11}\\
        & 4     & & 23.85 & 26.59 & 23.95 & 35.91 & \textbf{38.53}\\
        & 8     & & 26.23 & 26.46 & 24.78 & 37.70 & \textbf{39.26}\\
        & 16    & & 28.48 & 27.67 & 26.27 & 39.69 & \textbf{40.00}\\
\bottomrule
\end{tabular}}
\label{medical}
\end{table}

\begin{table}[t]
  \centering
  \caption{1-shot and 5-shot mean IoU (\%) on Pascal VOC. Backbone = ViT-B/16.}
  \label{tab:voc-seg}\scalebox{0.99}{
  \begin{tabular}{@{}lcc@{}}
    \toprule
    \textbf{Method} & \textbf{1-shot} & \textbf{5-shot} \\
    \midrule
    CoOp                    & 41.92 ± 0.24 & 52.37 ± 0.28 \\
    CuPL*+WiSE             & \textit{43.55 ± 0.22} & \textit{54.01 ± 0.25} \\
    \textbf{CAPEL (ours)}   & \textbf{45.28 ± 0.19} & \textbf{55.74 ± 0.21} \\
    \bottomrule
  \end{tabular}}
  \vspace{-10pt}
  \label{segtable}
\end{table}

\subsection{Generalization to Medical and Industrial Domains}
We further add medical and industrial benchmarks to complement the 11 natural-image datasets. Following BiomedCoOp\cite{koleilat2025biomedcoop}, we evaluate on the gastrointestinal endoscopy dataset \textsc{Kvasir}  \cite{pogorelov2017kvasir} and the X-ray dataset \textsc{KneeXray} \cite{chen2018knee}. CAPEL exceeds CLIP adaptation baselines on all shot settings (Table~\ref{medical}), except for 2-shot setting on Kvasir. For the industrial setting, following PromptAD\cite{li2024promptad}, we include the industrial anomaly-detection datasets \textsc{MVTec} and \textsc{VisA}, where CAPEL maintains strong performance under the standard evaluation protocols as shown in Table \ref{industy}. Overall, the results confirm that CAPEL generalises across application domains (natural, medical, and industrial images) without modifying the backbone or increasing inference cost.

\subsection{Generalization to Semantic Segmentation}
\begin{figure}[t!]
\begin{center}
\includegraphics[width=0.99\linewidth]{ 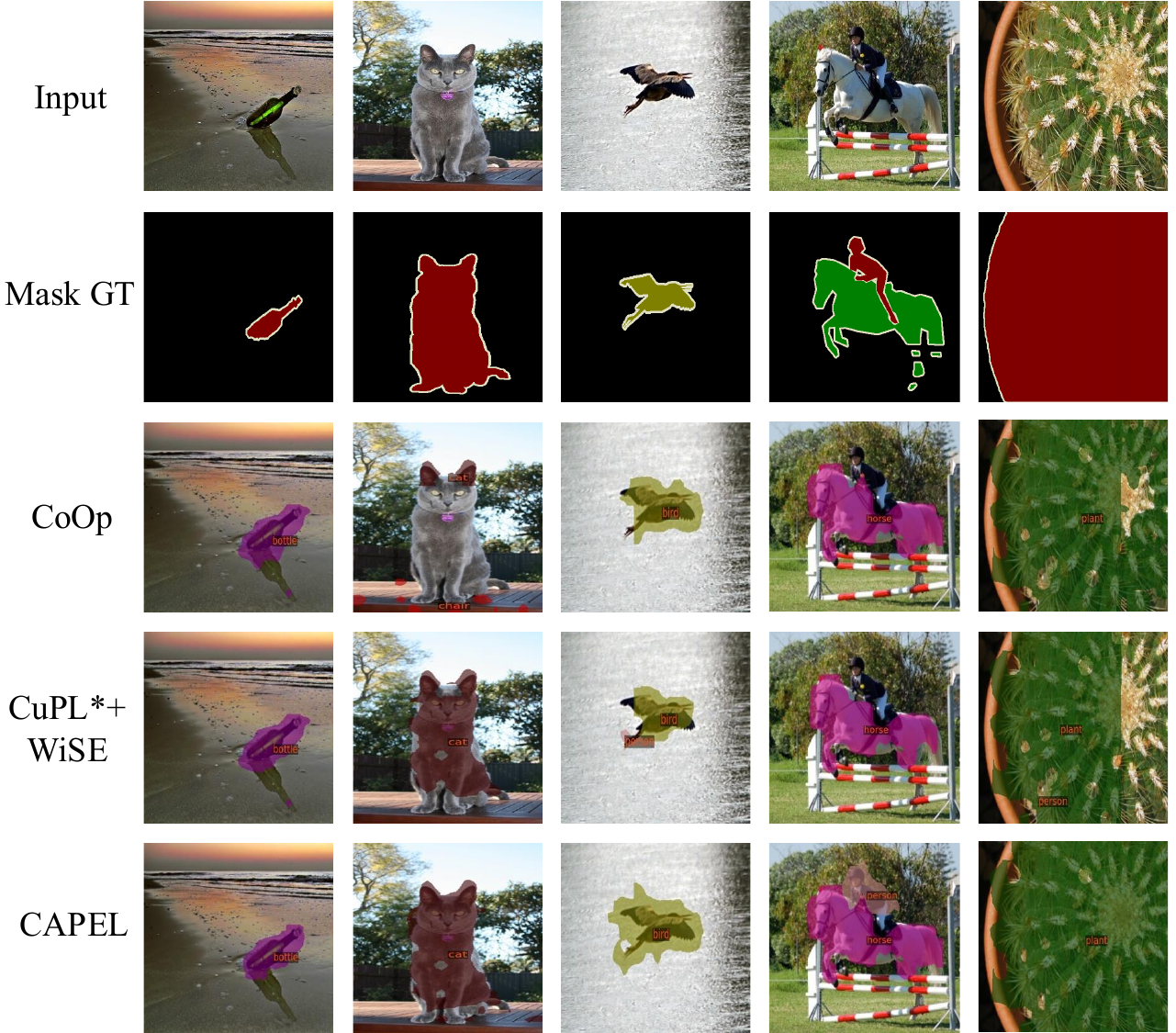}
\vspace{-10pt}
\end{center}
  \caption{Qualitative PASCAL VOC-2012 examples (val). Top to bottom: input image, ground-truth mask, CoOp, CuPL*+WiSE, and CAPEL results. CAPEL yields cleaner boundaries and fewer spurious regions under the same decoder and backbone.}
\label{segfig}
\vspace{-10pt}
\end{figure}

To evaluate CAPEL beyond classification, we conducted a proof-of-concept experiment on \textbf{few-shot semantic segmentation} as shown in Table \ref{segtable} and Figure \ref{segfig}.  
We adapted the publicly available \textsc{CLIPSeg} decoder \cite{luddecke2022image} and replaced its classifier layer with our \emph{logit-ensemble of $K$ sub-classifiers per category}. All hyper-parameters were kept identical to the classification setting. Although CAPEL is not specificially designed for segmentation tasks, it can be seen that CAPEL also performs superiorly in the segmentation task compared with the baseline methods. It shows the potential generality of the logits emsembling technique to downstream tasks. We have included the experimental results to the revised manuscript.

\subsection{Hyper-Parameter Analysis}
\begin{figure}[t]
    \centering
    \subfloat[Impact of the prompt number]
    {\includegraphics[width=1.7in]{ 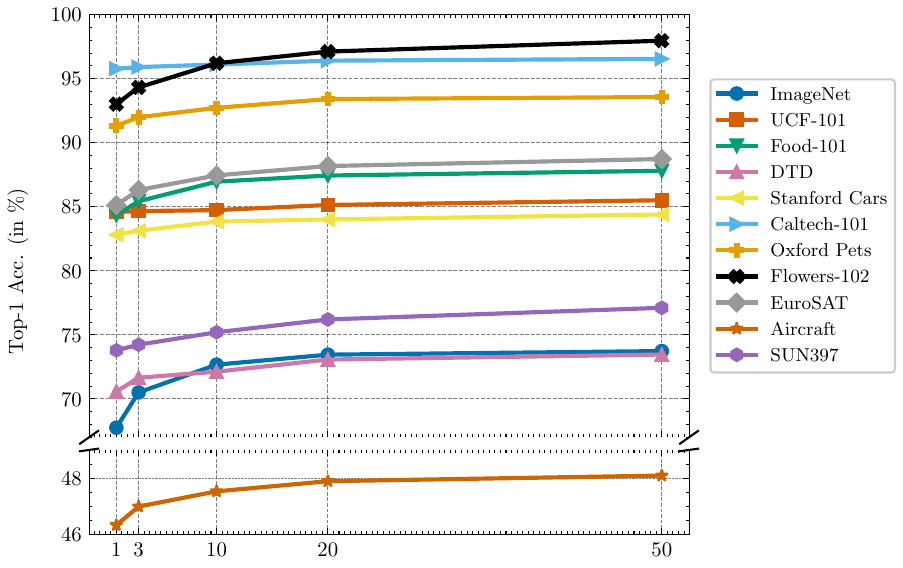} \vspace{-0pt}}
    \subfloat[Impact of $\ell_{pc}$]
    {\includegraphics[width=1.7in]{ 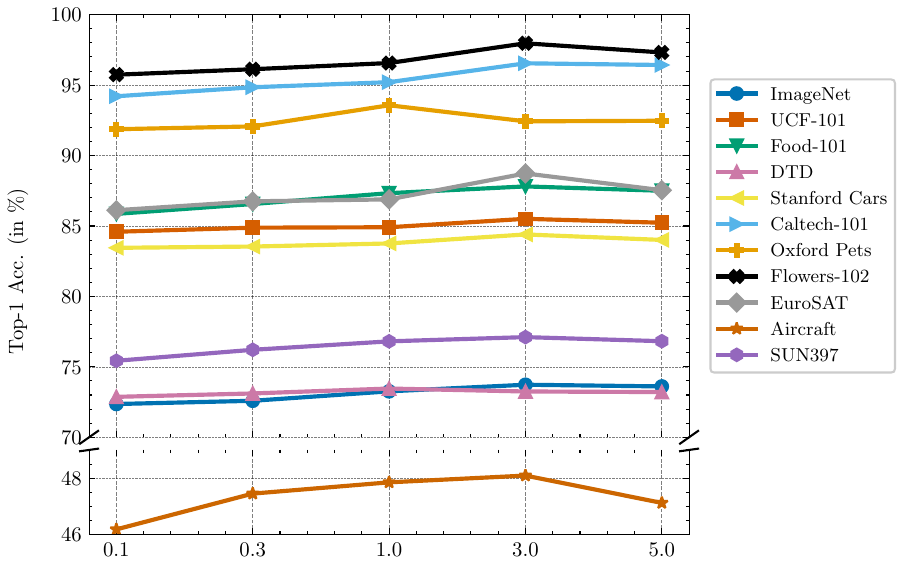} \vspace{-0pt}}\\
    \caption{Hyper-parameter Sensitivity on prompt number $K$ and the loss weight for cluster-preserving regularization $\ell_{pc}$. We show that using multiple sub-prompts for each class achieves better performance. We conduct experiments to check the hyper-parameter sensitivity of the coefficients that control $\ell_{pc}$, which achieve the best performance in 3.}\label{perturbation_vis}
    \label{hyper}
    \vspace{-5pt}
\end{figure}

To study the impact of the number of class sub-prompts, we respectively construct 1, 3, 10, 20, and 50 class sub-prompts for all 11 datasets. 
In Fig. \ref{perturbation_vis} (a), the lines represent the performance trend of CAPEL when increasing the sub-prompt number.
As we vary the number of prompts from 1 to 50, there is an obvious uptrend in few-shot performance.
The finding highlights our core motivation that a singular prompt cannot cover the entire variance space of a visual category, but employing a diverse range of sub-prompts is instrumental in encompassing the extensive variance. We also report the effects of varying the loss coefficients $\lambda$  for and $\ell_{pc}$. As shown in Fig. \ref{hyper} (b), $\lambda$ is set to 3 to achieve the best performance.

\begin{figure}[t]
  \centering
  \includegraphics[width=0.5\textwidth]{ 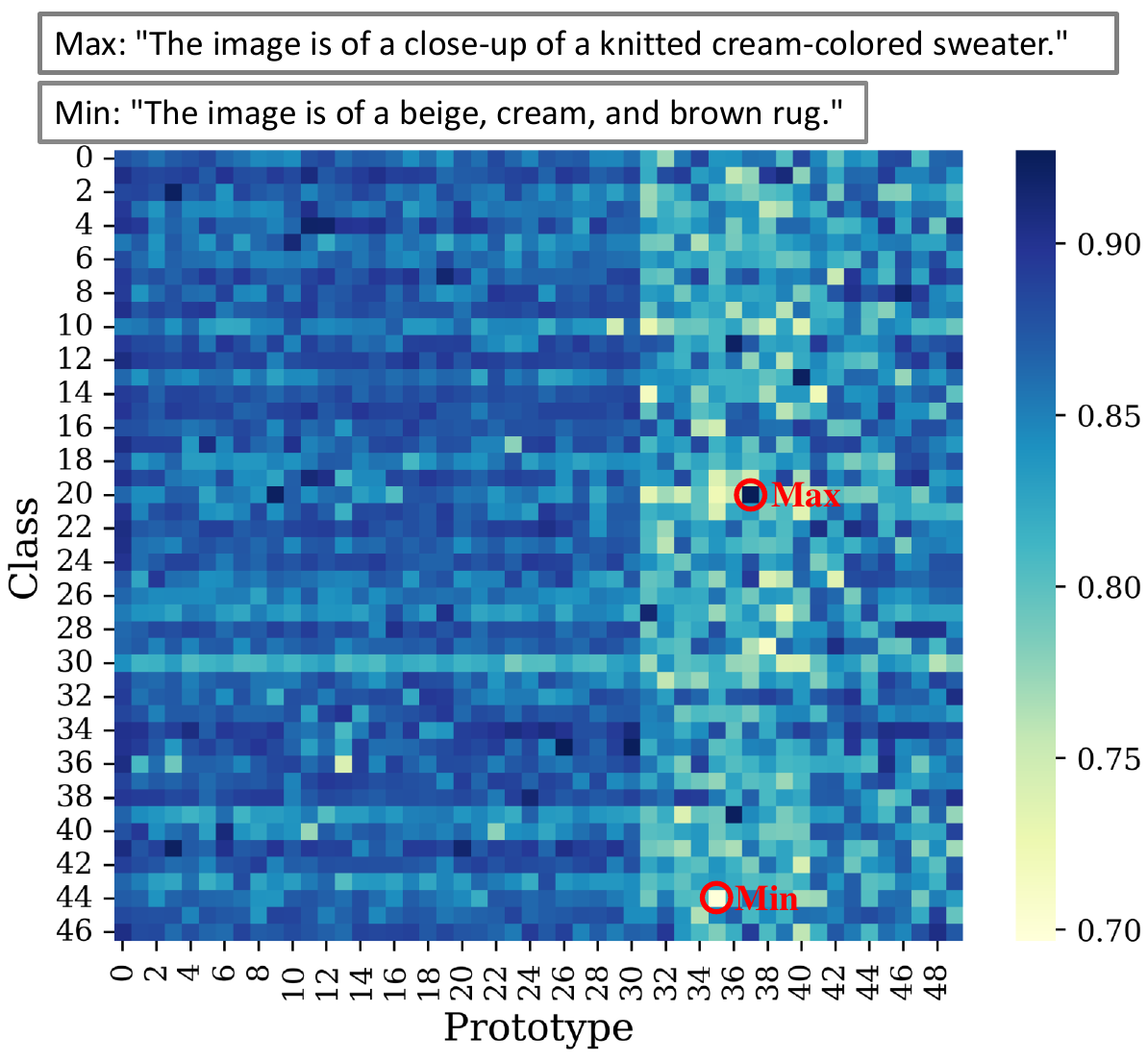}
  \caption{Heatmap visualization of the attention matrix on DTD. Lighter color represents less contribution of the final prediction, while darker color indicates high importance.}
  \label{fig:attention}
\end{figure}

\begin{figure*}[h!]
  \centering
  \includegraphics[width=0.95\textwidth]{ 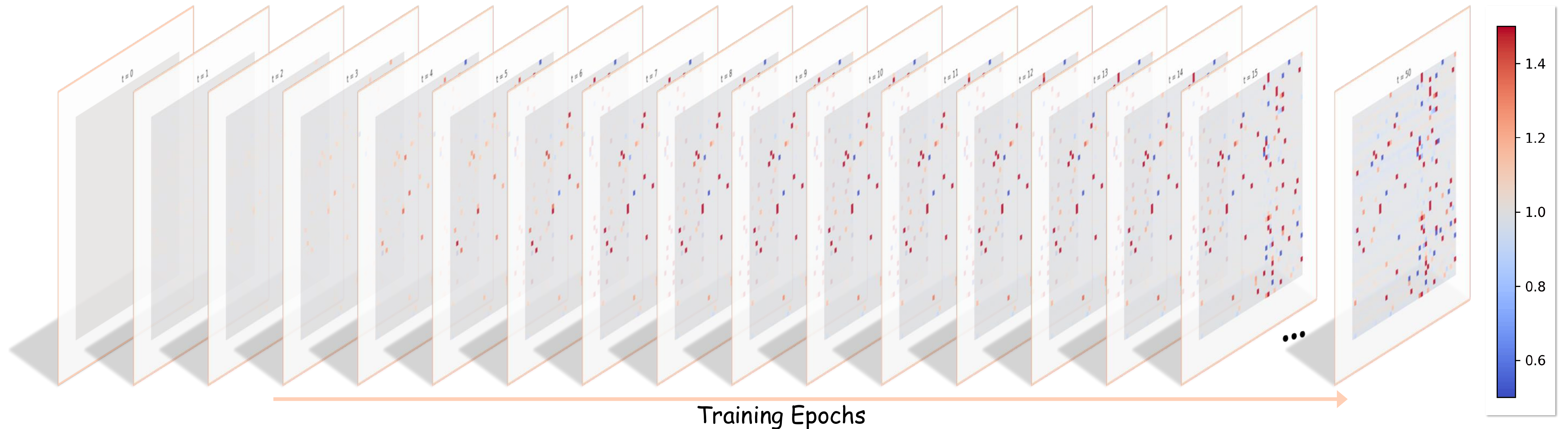}
  \caption{Attention weights visualization over epochs. The dynamics show the model picks up the important prompts while penalizing the noisy ones.}
  \label{fig:alpha-dynamics}
\end{figure*}

\begin{table*}[t]
\caption{Case study of prototype learning. We explain the learned prototypes by showing the most similar words between initialization and after fine-tuning. } 
\vspace{-10pt}
\begin{center}
\scalebox{0.76}{\begin{tabular}{ c | cc|cc|cc}
\toprule 
 & \multicolumn{2}{c}{Food101} & \multicolumn{2}{c}{OxfordFlowers}  & \multicolumn{2}{c}{DTD}  \\
\midrule
\#  & Before & After  & Before & After & Before & After \\
\midrule
1 & desserts(0.77)  & waffle(0.62)      & inflorescence(0.78)& inflorescence(0.67)          & interlanced(0.84) & layered(0.83)\\ 
2 & food(0.77)      & pastry(0.61)      & leaves(0.67)      & hellebore(0.55)               & layered(0.83)     & interlaced(0.82)\\
3 & pizza(0.77)     & dessert(0.61)     & foliage(0.67)     & fruit(0.54)                   & indented(0.83)    & stripe(0.82)\\
4 & pastry(0.77)    & cookie(0.60)      & wildflower(0.65)  & floral(0.54)                  & weathered(0.83)   & indented(0.82)\\
5 & fruit(0.77)     & cheesecake(0.60)  & habit(0.65)       & pansy(0.54)                   & intricately(0.83) & patterns(0.82)\\
6 & bread(0.76)     & cake(0.59)        & leaf(0.64)        & rose(0.53)                    & patterns(0.83)    & rippled(0.81)\\
7 & cake(0.76)      & potato(0.59)      & fruit(0.64)       & florets(0.52)                 & details(0.83)     & mixed(0.81)\\
8 & cookie(0.76)    & maple(0.59)       & lilies(0.63)      & bloom(0.52)                   & veined(0.83)      & stratified(0.81)\\
9 & pork(0.76)      & pudding(0.59)     & blooms(0.63)      & pistils(0.52)                 & tactile(0.83)     & styled(0.81)\\
10& pancake(0.75)   & square(0.59)      & succulent(0.63)   & tropical(0.52)                & rippled(0.82)     & weathered(0.81)\\
\bottomrule
\end{tabular}}
\end{center}
\label{words}
\vspace{-20pt}
\end{table*}

\subsection{Qualitative Study} 
\noindent \textbf{Visualization of the Impact of Cluster-Preserving Regularization.}~~We visualize the learned sub-classifiers with and without the cluster-preserving regularization. As shown in Fig. \ref{fig:vis}, the transparent round dots are the initial sub-prompts, and the asterisks are the learned sub-classifiers. We can see the learned classifiers in the left are as diverse as the original sub-classifiers before training, while the sub-classifiers without regularization tend to gather as a group. Thus, when solely relying on class-level cross-entropy loss, the cluster collapse issue arises. The proposed cluster-preserving can mitigate the issue and preserve the sub-classifier diversity.

\subsubsection{Adaptive Prompt Weighting Visualization}
To better understand the impact of adaptive prompt weighting, we visualize the learned attention matrix with a heatmap in Fig. \ref{fig:attention}. The weights of different prompts in each class vary significantly, ranging from around 0.68 to 0.93. We conduct a case study on the two prompts, which have the least and the maximum contribution to the corresponding categories. 
The sub-prompt with maximum attention is ``The image is of a close-up of a knitted cream-colored sweater.” Other prompts in the same category involve specific colors that cover a small portion of the visual samples, but ``cream-colored" is a more general color descriptor that may match more visual samples. Thus, a higher attention value is assigned. The sub-prompt of the least attention is ``The image is of a beige, cream, and brown rug.”, which involves a wrong category name ``Rug", rather than the ground-truth name ``Woven". This demonstrates that the adaptive prompt weighting technique can handle noisy prompts. Fig. \ref{fig:alpha-dynamics} shows the dynamics of the attention weights over epochs. The model gradually picks up the important prompts while penalizing the noisy ones.

\subsubsection{Interpreting the Learned Prompts}
To understand the changes to the prompts during fine-tuning,  we calculate the cosine similarity between sub-classifier weights and the clip textual features of the individual words from the prompts. We randomly selected three prmopts from Food101, OxfordFlowers, and DTD and analyzed the changes after fine-tuning. The original prompt examples are provided in Table \ref{prompts_sample}. The top 10 most similar words are listed together with the similarity in Table \ref{words}. Note that the class names and function words are removed from the corpus. We observe that the prototypes after training are more specific than before training. Particularly, in OxfordFlowers, all the top words after training are closer to the flowers rather than the leaves. 

\begin{table}[h!]
\scriptsize
\caption{{Prompt Samples of 11 Datasets}}
\scalebox{0.95}{
{\begin{tabularx}{0.5\textwidth}{l|l}
\toprule
Dataset          & Prompt Samples \\
\midrule
\multirow{3}{*}{\textit{ImageNet}}               
& There are many different types of military uniforms, but they all...\\ 
& The image is of a red and white mitten with a green background.  \\ 
& A graduation cap typically has a square or pyramid shape and is... 
\\
\midrule
\multirow{3}{*}{\textit{Caltech-101}}               
& A motorbike with two wheels and a seat. \\ 
& The image shows an airplane flying through the sky. \\ 
& You can identify a ant by its long, segmented body and its long...\\
\midrule
\multirow{3}{*}{\textit{Oxford Pets}}               
& The Russian Blue has a sleek and elegant coat of bluish-gray fur...\\ 
& The Scottish Terrier is a small, sturdy breed with a distinctive ...\\ 
& A hairless cat breed with wrinkled skin, large ears, and a slim ... \\

\midrule
\multirow{3}{*}{\textit{Stanford Cars }}               
& The 2007 Dodge Dakota Club Cab is a four-door truck with a ...\\ 
& A 1993 Geo Metro Convertible would look like a small, boxy ... \\ 
& The image is of a silver 2012 Dodge Charger Sedan. \\
\midrule
\multirow{3}{*}{\textit{Oxford Flowers}}               
& Cyclamen have heart-shaped leaves with vibrant flowers that  ... \\ 
& Frangipani is a stunning tropical flower with vibrant, star-shaped ... \\ 
& The sword lily features tall, slender stalks topped with vibrant ... \\
\midrule
\multirow{3}{*}{\textit{Food101}}               
& A baklava is a layered pastry made with nuts, honey, and phyllo dough.\\ 
& The image is of a beet salad with goat cheese, arugula, and pistachios. \\ 
& A bread pudding generally has a bread base with eggs, milk, and sugar...\\

\midrule
\multirow{3}{*}{\textit{FGVC Aircraft }}               
& A 737-300 aircraft operated by SouthWest Airlines takes off from the ... \\ 
& The image is of an aircraft 737-400 with the engines running. The plane ... \\ 
&  Bustling with activity, this Gulfstream IV is parked on the tarmac ...\\
\midrule
\multirow{3}{*}{\textit{SUN397}}               
& This image is of the Abbey of Saint-Denis, a large abbey located in the ... \\ 
& An image of an amusement arcade shows a large room with brightly lit  ... \\ 
& A ticket booth and information desk are visible in this shot of ...\\
\midrule
\multirow{3}{*}{\textit{DTD}}               
& A bubbly texture can look like small mountains with peaks that are ...\\ 
& A chequered texture is a texture that has a repeating pattern of ...\\ 
& The crystalline texture is characterized by having a distinct...\\
\midrule
\multirow{3}{*}{\textit{EuroSAT }}               
& A centered satellite image of a River displays areas of human ...\\ 
& A centered satellite image of Residential Buildings displays clusters... \\ 
& A centered satellite image of Forest displays clear linear patterns ... \\
\midrule
\multirow{3}{*}{\textit{UCF101 }}               
& A person is applying lipstick in the image.\\ 
& A person doing Baby Crawling looks like a human infant crawling...\\ 

& The person is doing a handstand on the balance beam. \\
\bottomrule
\end{tabularx}}}
\label{prompts_sample}
\vspace{-20pt}
\end{table}

\section{Conclusion, Limitations and Future Work}
In this paper, we introduce the Cluster-Aware Prompt Ensemble Learning (CAPEL) framework for vision-language models, which significantly enhances zero-shot CLIP performance by incorporating multiple prompts as class sub-prototypes. We have implemented a prototype competition technique to ensure that each prototype specializes in a distinct subset of the class. Furthermore, to address the inherent noise and flaws in these prompts, we have developed an adaptive attention mechanism. CAPEL has demonstrated consistent performance improvements across all 11 datasets and various tasks. Moreover, this paper offers comprehensive insights into the utility of class sub-prototypes, shedding light on prototype learning and the broader implications for vision-language models.

At the same time, several limitations remain. CAPEL relies on an external text generator to seed a diverse prompt bank. Although open-source LLMs are supported and prompts are released, access or licensing may be restrictive in some settings. Training-time overhead can grow with the number of prompts per class, and while pruning largely recovers throughput, the worst-case configuration is less efficient than single-prompt baselines. Gains taper as shot counts become very high, where strong feature-level methods already saturate. The conditional-entropy regularizer implicitly assumes that classes exhibit multiple latent visual clusters. When this assumption is weak, such as single-mode classes, the advantage of logit-space ensembling may diminish.

These observations suggest several avenues for future work. A first direction is greater adaptivity: learning the number of active prompts per class and exploring instance-conditioned routing that preserves fully batched inference. A second is removing external dependencies via lightweight, on-device prompt generators and active or curriculum strategies for curating prompt banks. Beyond this, integrating CAPEL with parameter-efficient tuning of the vision encoder, extending to detection/segmentation and video (including industrial anomaly localization and medical imaging), and analyzing when logit-space ensembling provably dominates feature-space averaging are promising next steps. We also plan to study continual and cross-domain adaptation, where CAPEL’s prunable heads and class-specific weights could support long-horizon learning under distribution shift.


\section*{Acknowledgement}
This work is supported by the Australian Research Council under the streams of Discovery Project (No. DP240101814).







\bibliographystyle{elsarticle-num}
\bibliography{bib}

\begin{thebibliography}{10}
\expandafter\ifx\csname url\endcsname\relax
  \def\url#1{\texttt{#1}}\fi
\expandafter\ifx\csname urlprefix\endcsname\relax\def\urlprefix{URL }\fi
\expandafter\ifx\csname href\endcsname\relax
  \def\href#1#2{#2} \def\path#1{#1}\fi

\bibitem{radford2021learning}
A.~Radford, J.~W. Kim, C.~Hallacy, A.~Ramesh, G.~Goh, S.~Agarwal, G.~Sastry, A.~Askell, P.~Mishkin, J.~Clark, et~al., Learning transferable visual models from natural language supervision, in: ICML, 2021, pp. 8748--8763.

\bibitem{jia2021scaling}
C.~Jia, Y.~Yang, Y.~Xia, Y.-T. Chen, Z.~Parekh, H.~Pham, Q.~Le, Y.-H. Sung, Z.~Li, T.~Duerig, Scaling up visual and vision-language representation learning with noisy text supervision, in: ICML, 2021, pp. 4904--4916.

\bibitem{pham2023combined}
H.~Pham, Z.~Dai, G.~Ghiasi, K.~Kawaguchi, H.~Liu, A.~W. Yu, J.~Yu, Y.-T. Chen, M.-T. Luong, Y.~Wu, et~al., Combined scaling for zero-shot transfer learning, Neurocomputing 555 (2023) 126658.

\bibitem{pratt2023does}
S.~Pratt, I.~Covert, R.~Liu, A.~Farhadi, What does a platypus look like? generating customized prompts for zero-shot image classification, in: ICCV, 2023, pp. 15691--15701.

\bibitem{wortsman2022robust}
M.~Wortsman, G.~Ilharco, J.~W. Kim, M.~Li, S.~Kornblith, R.~Roelofs, R.~G. Lopes, H.~Hajishirzi, A.~Farhadi, H.~Namkoong, et~al., Robust fine-tuning of zero-shot models, in: Proceedings of the IEEE/CVF conference on computer vision and pattern recognition, 2022, pp. 7959--7971.

\bibitem{allingham2023simple}
J.~U. Allingham, J.~Ren, M.~W. Dusenberry, X.~Gu, Y.~Cui, D.~Tran, J.~Z. Liu, B.~Lakshminarayanan, A simple zero-shot prompt weighting technique to improve prompt ensembling in text-image models, in: International Conference on Machine Learning, PMLR, 2023, pp. 547--568.

\bibitem{zhou2022learning}
K.~Zhou, J.~Yang, C.~C. Loy, Z.~Liu, Learning to prompt for vision-language models, IJCV (2022) 2337--2348.

\bibitem{gao2024clip}
P.~Gao, S.~Geng, R.~Zhang, T.~Ma, R.~Fang, Y.~Zhang, H.~Li, Y.~Qiao, Clip-adapter: Better vision-language models with feature adapters, IJCV 132~(2) (2024) 581--595.

\bibitem{devlin2018bert}
J.~Devlin, M.-W. Chang, K.~Lee, K.~Toutanova, Bert: Pre-training of deep bidirectional transformers for language understanding, arXiv:1810.04805 (2018).

\bibitem{radford2019language}
A.~Radford, J.~Wu, R.~Child, D.~Luan, D.~Amodei, I.~Sutskever, et~al., Language models are unsupervised multitask learners, OpenAI blog 1~(8) (2019) 9.

\bibitem{brown2020language}
T.~Brown, B.~Mann, N.~Ryder, M.~Subbiah, J.~D. Kaplan, P.~Dhariwal, A.~Neelakantan, P.~Shyam, G.~Sastry, A.~Askell, et~al., Language models are few-shot learners, NeurIPS 33 (2020) 1877--1901.

\bibitem{chen2020canzsl}
Z.~Chen, J.~Li, Y.~Luo, Z.~Huang, Y.~Yang, Canzsl: Cycle-consistent adversarial networks for zero-shot learning from natural language, in: IEEE/CVF Winter Conference on Applications of Computer Vision (WACV), 2020, pp. 874--883.

\bibitem{chen2020rethinking}
Z.~Chen, S.~Wang, J.~Li, Z.~Huang, Rethinking generative zero-shot learning: An ensemble learning perspective for recognising visual patches, in: Proceedings of the 28th ACM International Conference on Multimedia, 2020, pp. 3413--3421.

\bibitem{chen2021entropy}
Z.~Chen, Z.~Huang, J.~Li, Z.~Zhang, Entropy-based uncertainty calibration for generalized zero-shot learning, in: Australasian Database Conference 2021, 2021.

\bibitem{chen2021mitigating}
Z.~Chen, Y.~Luo, S.~Wang, R.~Qiu, J.~Li, Z.~Huang, Mitigating generation shifts for generalized zero-shot learning, in: Proceedings of the 28th ACM International Conference on Multimedia, 2021.

\bibitem{chen2021semantics}
Z.~Chen, Y.~Luo, R.~Qiu, S.~Wang, Z.~Huang, J.~Li, Z.~Zhang, Semantics disentangling for generalized zero-shot learning, in: IEEE/CVF International Conference on Computer Vision (ICCV), 2021.

\bibitem{su2022distinguishing}
H.~Su, J.~Li, Z.~Chen, L.~Zhu, K.~Lu, Distinguishing unseen from seen for generalized zero-shot learning, in: IEEE/CVF Conference on Computer Vision and Pattern Recognition (CVPR), 2022.

\bibitem{chen2022federated}
Z.~Chen, Y.~Luo, S.~Wang, J.~Li, Z.~Huang, Federated zero-shot learning for visual recognition, arXiv preprint arXiv:2209.01994 (2022).

\bibitem{chen2022gsmflow}
Z.~Chen, Y.~Luo, S.~Wang, J.~Li, Z.~Huang, Gsmflow: Generation shifts mitigating flow for generalized zero-shot learning, IEEE Transactions on Multimedia (2022).

\bibitem{chen2023zero}
Z.~Chen, P.~Zhang, J.~Li, S.~Wang, Z.~Huang, Zero-shot learning by harnessing adversarial samples, in: Proceedings of the 31th ACM International Conference on Multimedia 2023, 2023.

\bibitem{guo2024fine}
J.~Guo, Z.~Rao, Z.~Chen, J.~Zhou, D.~Tao, Fine-grained zero-shot learning: Advances, challenges, and prospects, arXiv preprint arXiv:2401.17766 (2024).

\bibitem{guo2024element}
J.~Guo, Z.~Rao, Z.~Chen, S.~Guo, J.~Zhou, D.~Tao, On the element-wise representation and reasoning in zero-shot image recognition: A systematic survey, arXiv preprint arXiv:2408.04879 (2024).

\bibitem{chen2025svip}
Z.~Chen, Z.~Zhao, J.~Guo, J.~Li, Z.~Huang, Svip: Semantically contextualized visual patches for zero-shot learning, in: ICCV2025, 2025.

\bibitem{su2019vl}
W.~Su, X.~Zhu, Y.~Cao, B.~Li, L.~Lu, F.~Wei, J.~Dai, Vl-bert: Pre-training of generic visual-linguistic representations, arXiv:1908.08530 (2019).

\bibitem{tan2019lxmert}
H.~Tan, M.~Bansal, Lxmert: Learning cross-modality encoder representations from transformers, arXiv:1908.07490 (2019).

\bibitem{yuan2021florence}
L.~Yuan, D.~Chen, Y.-L. Chen, N.~Codella, X.~Dai, J.~Gao, H.~Hu, X.~Huang, B.~Li, C.~Li, et~al., Florence: A new foundation model for computer vision, arXiv:2111.11432 (2021).

\bibitem{yang2023effective}
X.~Yang, F.~Liu, G.~Lin, Effective end-to-end vision language pretraining with semantic visual loss, IEEE Transactions on Multimedia 25 (2023) 8408--8417.

\bibitem{gao2021clip}
P.~Gao, S.~Geng, R.~Zhang, T.~Ma, R.~Fang, Y.~Zhang, H.~Li, Y.~Qiao, Clip-adapter: Better vision-language models with feature adapters, arXiv:2110.04544 (2021).

\bibitem{kim2021adapt}
K.~Kim, M.~Laskin, I.~Mordatch, D.~Pathak, How to adapt your large-scale vision-and-language model (2021).

\bibitem{zhang2021tip}
R.~Zhang, R.~Fang, W.~Zhang, P.~Gao, K.~Li, J.~Dai, Y.~Qiao, H.~Li, Tip-adapter: Training-free clip-adapter for better vision-language modeling, arXiv:2111.03930 (2021).

\bibitem{xing2023dual}
Y.~Xing, Q.~Wu, D.~Cheng, S.~Zhang, G.~Liang, P.~Wang, Y.~Zhang, Dual modality prompt tuning for vision-language pre-trained model, IEEE Transactions on Multimedia (2023).

\bibitem{wei2024benchmarking}
T.~Wei, Z.~Chen, Z.~Huang, X.~Yu, Benchmarking in-the-wild multimodal disease recognition and a versatile baseline, in: ACM Multimedia 2024, 2024.

\bibitem{wei2024snap}
T.~Wei, Z.~Chen, X.~Yu, Snap and diagnose: An advanced multimodal retrieval system for identifying plant diseases in the wild, in: Proceedings of the 6th ACM International Conference on Multimedia in Asia, 2024, pp. 1--3.

\bibitem{feng2022promptdet}
C.~Feng, Y.~Zhong, Z.~Jie, X.~Chu, H.~Ren, X.~Wei, W.~Xie, L.~Ma, Promptdet: Towards open-vocabulary detection using uncurated images, in: ECCV, Springer, 2022, pp. 701--717.

\bibitem{gu2021open}
X.~Gu, T.-Y. Lin, W.~Kuo, Y.~Cui, Open-vocabulary object detection via vision and language knowledge distillation, arXiv:2104.13921 (2021).

\bibitem{zhang2025cross}
G.~Zhang, S.~Kan, L.~Shi, W.~Xu, G.~An, Y.~Cen, Cross-scene visual context parsing with large vision-language model, Pattern Recognition (2025) 111641.

\bibitem{liu2025vision}
X.~Liu, W.~Gong, X.~Chen, Z.~Li, Y.~Liu, L.~Wang, Q.~Liu, X.~Sun, X.~Liu, X.~Chen, et~al., Vision-language foundation model for generalizable nasal disease diagnosis using unlabeled endoscopic records, Pattern Recognition (2025) 111646.

\bibitem{cai2025prompt}
S.~Cai, X.~Liu, J.~Yuan, Q.~Zhou, Prompt-ladder: Memory-efficient prompt tuning for vision-language models on edge devices, Pattern Recognition 163 (2025) 111460.

\bibitem{dong2024cluster}
M.~Dong, F.~Li, Z.~Li, X.~Liu, Cluster prototype earth mover’s distance adapters and alignment-guided prompt learning for vision--language models, Pattern Recognition 156 (2024) 110861.

\bibitem{han2024f}
B.~Han, X.~Jiang, Z.~Fang, H.~Fujita, Y.~Gao, F-scp: An automatic prompt generation method for specific classes based on visual language pre-training models, Pattern Recognition 147 (2024) 110096.

\bibitem{huang2024joint}
R.~Huang, X.~Pan, H.~Zheng, H.~Jiang, Z.~Xie, C.~Wu, S.~Song, G.~Huang, Joint representation learning for text and 3d point cloud, Pattern Recognition 147 (2024) 110086.

\bibitem{lim2024dipex}
J.~S. Lim, Z.~Chen, Z.~Chen, M.~Baktashmotlagh, X.~Yu, Z.~Huang, Y.~Luo, Dipex: Dispersing prompt expansion for class-agnostic object detection, NeurIPS2024 (2024).

\bibitem{zhang2024ta}
W.~Zhang, Y.~Zhang, Y.~Deng, W.~Zhang, J.~Lin, B.~Huang, J.~Zhang, W.~Yu, Ta-adapter: Enhancing few-shot clip with task-aware encoders, Pattern Recognition 153 (2024) 110559.

\bibitem{zhang2022tip}
R.~Zhang, W.~Zhang, R.~Fang, P.~Gao, K.~Li, J.~Dai, Y.~Qiao, H.~Li, Tip-adapter: Training-free adaption of clip for few-shot classification, in: ECCV, Springer, 2022, pp. 493--510.

\bibitem{shazeer2017outrageously}
N.~Shazeer, A.~Mirhoseini, K.~Maziarz, A.~Davis, Q.~Le, G.~Hinton, J.~Dean, Outrageously large neural networks: The sparsely-gated mixture-of-experts layer, in: International Conference on Learning Representations, 2017.

\bibitem{fedus2022switch}
W.~Fedus, B.~Zoph, N.~Shazeer, Switch transformers: Scaling to trillion parameter models with simple and efficient sparsity, Journal of Machine Learning Research 23~(120) (2022) 1--39.

\bibitem{lu2022prompt}
Y.~Lu, J.~Liu, Y.~Zhang, Y.~Liu, X.~Tian, Prompt distribution learning, in: CVPR, 2022, pp. 5206--5215.

\bibitem{khattak2023maple}
M.~U. Khattak, H.~Rasheed, M.~Maaz, S.~Khan, F.~S. Khan, Maple: Multi-modal prompt learning, in: CVPR, 2023, pp. 19113--19122.

\bibitem{zhu2023prompt}
B.~Zhu, Y.~Niu, Y.~Han, Y.~Wu, H.~Zhang, Prompt-aligned gradient for prompt tuning, in: ICCV, 2023, pp. 15659--15669.

\bibitem{yao2024tcp}
H.~Yao, R.~Zhang, C.~Xu, Tcp: Textual-based class-aware prompt tuning for visual-language model, in: Proceedings of the IEEE/CVF Conference on Computer Vision and Pattern Recognition, 2024, pp. 23438--23448.

\bibitem{xiao2010sun}
J.~Xiao, J.~Hays, K.~A. Ehinger, A.~Oliva, A.~Torralba, Sun database: Large-scale scene recognition from abbey to zoo, in: CVPR, 2010, pp. 3485--3492.

\bibitem{soomro2012ucf101}
K.~Soomro, A.~R. Zamir, M.~Shah, Ucf101: A dataset of 101 human actions classes from videos in the wild, arXiv:1212.0402 (2012).

\bibitem{helber2019eurosat}
P.~Helber, B.~Bischke, A.~Dengel, D.~Borth, Eurosat: A novel dataset and deep learning benchmark for land use and land cover classification, IEEE GRSS (2019) 2217--2226.

\bibitem{cimpoi2014describing}
M.~Cimpoi, S.~Maji, I.~Kokkinos, S.~Mohamed, A.~Vedaldi, Describing textures in the wild, in: CVPR, 2014, pp. 3606--3613.

\bibitem{deng2009imagenet}
J.~Deng, W.~Dong, R.~Socher, L.-J. Li, K.~Li, L.~Fei-Fei, Imagenet: A large-scale hierarchical image database, in: CVPR, 2009, pp. 248--255.

\bibitem{fei2004learning}
L.~Fei-Fei, R.~Fergus, P.~Perona, Learning generative visual models from few training examples: An incremental bayesian approach tested on 101 object categories, in: CVPRW, 2004, pp. 178--178.

\bibitem{parkhi2012cats}
O.~M. Parkhi, A.~Vedaldi, A.~Zisserman, C.~Jawahar, Cats and dogs, in: CVPR, 2012, pp. 3498--3505.

\bibitem{krause20133d}
J.~Krause, M.~Stark, J.~Deng, L.~Fei-Fei, 3d object representations for fine-grained categorization, in: ICCVW, 2013, pp. 554--561.

\bibitem{nilsback2008automated}
M.-E. Nilsback, A.~Zisserman, Automated flower classification over a large number of classes, in: ICVGIP, 2008, pp. 722--729.

\bibitem{bossard2014food}
L.~Bossard, M.~Guillaumin, L.~Van~Gool, Food-101--mining discriminative components with random forests, in: ECCV, Springer, 2014, pp. 446--461.

\bibitem{maji2013fine}
S.~Maji, E.~Rahtu, J.~Kannala, M.~Blaschko, A.~Vedaldi, Fine-grained visual classification of aircraft, arXiv:1306.5151 (2013).

\bibitem{recht2019imagenet}
B.~Recht, R.~Roelofs, L.~Schmidt, V.~Shankar, Do imagenet classifiers generalize to imagenet?, in: ICML, 2019, pp. 5389--5400.

\bibitem{wang2019learning}
H.~Wang, S.~Ge, Z.~Lipton, E.~P. Xing, Learning robust global representations by penalizing local predictive power, NeurIPS (2019).

\bibitem{hendrycks2021natural}
D.~Hendrycks, K.~Zhao, S.~Basart, J.~Steinhardt, D.~Song, Natural adversarial examples, in: CVPR, 2021, pp. 15262--15271.

\bibitem{hendrycks2021many}
D.~Hendrycks, S.~Basart, N.~Mu, S.~Kadavath, F.~Wang, E.~Dorundo, R.~Desai, T.~Zhu, S.~Parajuli, M.~Guo, et~al., The many faces of robustness: A critical analysis of out-of-distribution generalization, in: ICCV, 2021, pp. 8340--8349.

\bibitem{glorot2010understanding}
X.~Glorot, Y.~Bengio, Understanding the difficulty of training deep feedforward neural networks, in: AISTATS, JMLR Workshop and Conference Proceedings, 2010, pp. 249--256.

\bibitem{he2015delving}
K.~He, X.~Zhang, S.~Ren, J.~Sun, Delving deep into rectifiers: Surpassing human-level performance on imagenet classification, in: ICCV, 2015, pp. 1026--1034.

\bibitem{cohen2020sub}
N.~Cohen, Y.~Hoshen, Sub-image anomaly detection with deep pyramid correspondences, arXiv preprint arXiv:2005.02357 (2020).

\bibitem{defard2021padim}
T.~Defard, A.~Setkov, A.~Loesch, R.~Audigier, Padim: a patch distribution modeling framework for anomaly detection and localization, in: International conference on pattern recognition, Springer, 2021, pp. 475--489.

\bibitem{roth2022towards}
K.~Roth, L.~Pemula, J.~Zepeda, B.~Sch{\"o}lkopf, T.~Brox, P.~Gehler, Towards total recall in industrial anomaly detection, in: Proceedings of the IEEE/CVF conference on computer vision and pattern recognition, 2022, pp. 14318--14328.

\bibitem{jeong2023winclip}
J.~Jeong, Y.~Zou, T.~Kim, D.~Zhang, A.~Ravichandran, O.~Dabeer, Winclip: Zero-/few-shot anomaly classification and segmentation, in: Proceedings of the IEEE/CVF Conference on Computer Vision and Pattern Recognition, 2023, pp. 19606--19616.

\bibitem{tamura2023random}
M.~Tamura, Random word data augmentation with clip for zero-shot anomaly detection, arXiv preprint arXiv:2308.11119 (2023).

\bibitem{fang2023fastrecon}
Z.~Fang, X.~Wang, H.~Li, J.~Liu, Q.~Hu, J.~Xiao, Fastrecon: Few-shot industrial anomaly detection via fast feature reconstruction, in: Proceedings of the IEEE/CVF International Conference on Computer Vision, 2023, pp. 17481--17490.

\bibitem{li2024promptad}
X.~Li, Z.~Zhang, X.~Tan, C.~Chen, Y.~Qu, Y.~Xie, L.~Ma, Promptad: Learning prompts with only normal samples for few-shot anomaly detection, in: Proceedings of the IEEE/CVF Conference on Computer Vision and Pattern Recognition, 2024, pp. 16838--16848.

\bibitem{achiam2023gpt}
J.~Achiam, S.~Adler, S.~Agarwal, L.~Ahmad, I.~Akkaya, F.~L. Aleman, D.~Almeida, J.~Altenschmidt, S.~Altman, S.~Anadkat, et~al., Gpt-4 technical report, arXiv preprint arXiv:2303.08774 (2023).

\bibitem{meta2025llama4scout}
{Meta AI}, Llama 4 scout 17b-16e, \url{https://huggingface.co/meta-llama/Llama-4-Scout-17B-16E}, model card (open-weights release) (2025).

\bibitem{team2024qwen2}
Q.~Team, Qwen2 technical report, arXiv preprint arXiv:2407.10671 (2024).

\bibitem{deepmind_gemini25pro_2025}
{Google DeepMind}, Gemini 2.5 pro, \url{https://deepmind.google/models/gemini/pro/}, official model overview (2025).

\bibitem{schuhmann2022laion}
C.~Schuhmann, R.~Beaumont, R.~Vencu, C.~Gordon, R.~Wightman, M.~Cherti, T.~Coombes, A.~Katta, C.~Mullis, M.~Wortsman, et~al., Laion-5b: An open large-scale dataset for training next generation image-text models, NeurIPS (2022) 25278--25294.

\bibitem{li2022blip}
J.~Li, D.~Li, C.~Xiong, S.~Hoi, Blip: Bootstrapping language-image pre-training for unified vision-language understanding and generation, in: ICML, 2022, pp. 12888--12900.

\bibitem{koleilat2025biomedcoop}
T.~Koleilat, H.~Asgariandehkordi, H.~Rivaz, Y.~Xiao, Biomedcoop: Learning to prompt for biomedical vision-language models, in: Proceedings of the Computer Vision and Pattern Recognition Conference, 2025, pp. 14766--14776.

\bibitem{pogorelov2017kvasir}
K.~Pogorelov, K.~R. Randel, C.~Griwodz, S.~L. Eskeland, T.~de~Lange, D.~Johansen, C.~Spampinato, D.-T. Dang-Nguyen, M.~Lux, P.~T. Schmidt, et~al., Kvasir: A multi-class image dataset for computer aided gastrointestinal disease detection, in: Proceedings of the 8th ACM on Multimedia Systems Conference, 2017, pp. 164--169.

\bibitem{chen2018knee}
P.~Chen, Knee osteoarthritis severity grading dataset, Mendeley Data 1~(10.17632) (2018) 30784984.

\bibitem{luddecke2022image}
T.~L{\"u}ddecke, A.~Ecker, Image segmentation using text and image prompts, in: CVPR, 2022, pp. 7086--7096.

\end{thebibliography}

\end{document}